\def\year{2021}\relax
\documentclass[letterpaper]{article} % DO NOT CHANGE THIS
\usepackage{aaai21}  % DO NOT CHANGE THIS
\usepackage{times}  % DO NOT CHANGE THIS
\usepackage{helvet} % DO NOT CHANGE THIS
\usepackage{courier}  % DO NOT CHANGE THIS
\usepackage[hyphens]{url}  % DO NOT CHANGE THIS
\usepackage{graphicx} % DO NOT CHANGE THIS
\usepackage{natbib}  % DO NOT CHANGE THIS AND DO NOT ADD ANY OPTIONS TO IT
\usepackage{caption} % DO NOT CHANGE THIS AND DO NOT ADD ANY OPTIONS TO IT
\usepackage{booktabs}       % professional-quality tables
\usepackage{amsfonts}       % blackboard math symbols
\usepackage{nicefrac}       % compact symbols for 1/2, etc.
\usepackage{microtype}      % microtypography
\usepackage{amsmath}
\usepackage{amsfonts}
\usepackage{amssymb}
\usepackage{bm}
\usepackage{bbm}
\usepackage{graphicx}
\usepackage{multirow}

\usepackage{tikz}
\usetikzlibrary{decorations.pathmorphing}
\usetikzlibrary{arrows,shapes,decorations,automata,backgrounds,petri,calc}

\definecolor{labelcolor}{HTML}{3d3d5c}

\newtheorem{remark}{Remark}
\newtheorem{example}{Example}

\newcommand{\mbf}[1]{\mathbf{#1}}
\newcommand{\mcl}[1]{\mathcal{#1}}
\newcommand{\mbb}[1]{\mathbb{#1}}
\DeclareMathOperator*{\argmax}{argmax}

\newcommand{\citesupp}[1]{Appendix~\ref{#1}}

\title{Top-$k$ Ranking Bayesian Optimization}
\author {
        Quoc Phong Nguyen,\textsuperscript{\rm 1}
        Sebastian Tay,\textsuperscript{\rm 1}
        Bryan Kian Hsiang Low,\textsuperscript{\rm 1}
        Patrick Jaillet\textsuperscript{\rm 2} \\
}
\affiliations {
    \textsuperscript{\rm 1}Dept. of Computer Science, National University of Singapore, Republic of Singapore\\
    \textsuperscript{\rm 2}Dept. of Electrical Engineering and Computer Science, MIT, USA\\
    qphong@comp.nus.edu.sg, sebastian.tay@u.nus.edu, lowkh@comp.nus.edu.sg, jaillet@mit.edu
}

\begin{document}

\maketitle

\begin{abstract}
This paper presents a novel approach to \emph{{top-$k$} ranking Bayesian optimization} (\mbox{top-$k$} ranking BO) which is a practical and significant generalization of  preferential BO to handle top-$k$ ranking and tie/indifference observations.
We first design a surrogate model that is not only capable of catering to the above observations, but is also supported by a classic random utility model.
Another equally important contribution is the introduction of the first information-theoretic acquisition function in BO with preferential observation called \emph{multinomial predictive entropy search} (MPES) which is flexible in handling these observations and optimized for all inputs of a query jointly. 
MPES possesses superior performance compared with existing acquisition functions that select the inputs of a query one at a time greedily. We empirically evaluate the performance of MPES using several synthetic benchmark functions,  CIFAR-$10$ dataset, and  SUSHI preference dataset.
\end{abstract}

\section{Introduction}
\label{sec:intro}
\emph{Bayesian optimization} (BO) is an efficient approach to optimize expensive-to-evaluate black-box objective functions (i.e., possibly noisy, non-convex, and/or with no closed-form expression/derivative)  \citep{brochu10tut}.
In practice, direct access to function evaluations may not be always possible \citep{gonzalez2017preferential}. For example, a diner tasting two different dishes (i.e., inputs) can easily tell which dish he/she prefers, but it is relatively difficult for the diner to articulate a rigorous numeric value representing the taste of each dish (i.e., a function).
The difficulty in providing such a value arises from the need to taste all possible dishes (i.e., inputs) and assess the difference in the flavors of these dishes.
On the other hand, specifying the preference between dishes (i.e., inputs) is so natural that it becomes our daily dining routine.
Also, an inherent property of our preference, which should be built into the model, is our inability to elucidate the preference between very similar choices (i.e., indifference or a tie). For example, it is hardly possible for us to tell the difference between a cup of coffee with $20\%$ of sugar and another with $21\%$ of sugar.

To boost the practicality of BO, recent works on \emph{preferential BO} \citep{dewancker18,gonzalez2017preferential} have attempted to replace direct (but noisy) function evaluations/values with noisy preferences between inputs. 
In particular, given two inputs, the higher the objective function value at an input is, the more likely it is preferred.
The observation in these works is limited to a preference between a pair of inputs, i.e., a \emph{pairwise preference} \citep{dewancker18,gonzalez2017preferential}.
Furthermore, inputs in the pair are searched one after the other.
Ideally, we would like to search for the input pair jointly as the function values at these inputs are correlated.

Regarding the model, the work of \citet{gonzalez2017preferential} directly applies a \emph{Gaussian process} (GP) to model a latent preference function whose input is a pair of the objective function's inputs.
This model suffers from $3$ disadvantages: The input dimension of the GP is twice that of the objective function, the objective function is not modeled directly, and ties are not modeled.
The second drawback leads to the use of a computationally expensive soft-Copeland score to estimate the maximizer of the objective function. On the other hand, the work of \citet{dewancker18} employs the generalized Bradley-Terry model. Yet, it suffers from a crude mean field approximation which implies that the posterior beliefs of the function values at different inputs are independent from one another.
Note that there are several works with preference-based observations in bandit literature \citep{busa2018preference}. The most relevant work to preferential BO is the work of \citet{sui2017multi} where there is an infinite number of dependent arms modeled with a GP. 
However, it does not have a regret analysis like the other bandit algorithms. Besides, a probability density is modeled with a GP which allows negative values.

This paper presents an approach that can resolve both existing issues on the model and the acquisition function.
Our model is inspired from the multinomial logit model and its generalization to rankings and ties.
Combining with a GP, our model can be interpreted as a GP regression model with an i.i.d.~Gumbel noise.
As the GP directly models the underlying objective function, the maximizer of the objective function can be estimated with the maximizer of the GP posterior mean function like in the conventional BO, which is less computationally intensive than the soft-Copeland score in \cite{gonzalez2017preferential}.
Our model is capable of handling the observation as a ranking of the top-$k$ inputs in a finite set of inputs, i.e., \emph{top-$k$ rankings} (Sec.~\ref{subsec:topk}), and the possibility of a tie/indifference in the observation (Sec.~\ref{subsec:tie}).
The former subsumes the pairwise preference (i.e., a top-$1$ ranking of $2$ inputs) in the existing works \citep{dewancker18,gonzalez2017preferential}.
We call this generalized problem the \emph{top-$k$ ranking Bayesian optimization} (\mbox{top-$k$} BO).
Although our GP model has a non-Gaussian likelihood, it can be trained with \emph{variational inference} (Sec.~\ref{sec:vi}).

While information-theoretic acquisition functions \cite{es,pes,wang17mes} have been investigated extensively in the conventional BO and demonstrated promising performance, such a principled acquisition function, to the best of our knowledge, has not been explored in BO with preferential observation.
Therefore, to efficiently exploit the posterior belief provided by our model, we derive the first information-theoretic acquistion function for BO with preferential observation that is capable of handling different types of observation introduced in this paper. It maximizes the information gain on the maximizer of the objective function through observing the top-$k$ ranking observation, which we call \emph{multinomial predictive entropy search} (MPES) (Sec.~\ref{sec:mpes}).  
Apart from the fact that MPES is rooted in information theory, it can jointly search for all inputs of the query, which differs from existing acquisition functions for preferential BO \citep{dewancker18,gonzalez2017preferential} that search for inputs of a query one at a time greedily. We empirically evaluate our model and our acquisition function with different types of observation using several synthetic benchmark functions, CIFAR-$10$ dataset, and SUSHI preference dataset (Sec.~\ref{sec:experiment}).
\section{Multinomial Logit Model and\\ Top-$k$ Ranking Observations}
\label{sec:mlm}
Let $f: \mcl{X} \rightarrow \mbb{R}$ be an unknown objective function defined on a bounded input domain $\mcl{X} \subset \mbb{R}^d$.
A function evaluation/value at an input $\mbf{x} \in \mcl{X}$ is denoted as $f_{\mbf{x}} \in \mbb{R}$.
The goal is to search for the maximizer $\argmax_{\mbf{x} \in \mcl{X}} f_{\mbf{x}}$ by observing preferences between different inputs, i.e., observing the choice between different inputs based on noisy evaluation of $f$ at these inputs.
The noisy evaluation of $f$ can be viewed as the \emph{utility function} $u$ decomposed into $u_{\mbf{x}} \triangleq f_{\mbf{x}} + \epsilon_{\mbf{x}}$ where $\epsilon_{\mbf{x}}$ is a random noise.
This noise represents unknown factors that affect the preference but are not captured in our objective function $f_{\mbf{x}}$, which is a practical consideration. Recall our dining example in Sec.~\ref{sec:intro}, the objective function may not capture the effect of the food temperature or the diner's hunger on the preference of the dish.
Following the \emph{random utility model} \citep{marschak59}, an input $\mbf{x}$ is preferred over another input $\mbf{x}'$  (i.e., denoted as $\mbf{x} \succ \mbf{x}'$) if the difference $u_{\mbf{x}} - u_{\mbf{x}'}$ in the unknown utility function values at $\mbf{x}$ and $\mbf{x}'$ is at least a threshold $\delta \ge 0$. In other words, $p(\mbf{x} \succ \mbf{x}') = p(u_{\mbf{x}} - u_{\mbf{x}'} \ge \delta)$.  The threshold $\delta$ enables the possibility of a tie between $2$ inputs, which reflects the real-world scenario where one is indifferent between $\mbf{x}$ and $\mbf{x}'$ due to their similar utility values, i.e., $u_{\mbf{x}}$ and $u_{\mbf{x}'}$ are not sufficiently far apart. 
It subsumes an extreme case of $\delta = 0$, i.e., there is no tie in the observation such as in  \cite{gonzalez2017preferential}.

\nocite{train2009discrete}

To specify the noise $\epsilon_{\mbf{x}}$, a common approach in the literature of preference learning with GP \cite{chu2005preference,gonzalez2017preferential} is to assume a Gaussian noise. 
However, it is difficult to extend such a model of pairwise preferences to rankings, as explained in \citesupp{app:rankpairwise}.
On the other hand, we present a refreshing approach of modeling $\epsilon_{\mbf{x}}$ as a Gumbel noise such that we can leverage the well-established multinomial logit model \cite{mcfadden1974measurement} to enable rankings and ties in BO.

Under the Gumbel noise and given the objective function values, the probability that an input $\mbf{x}$ is preferred over a finite set $\mcl{C} \subset \mcl{X}$ of inputs (i.e., may or may not include $\mbf{x}$) has a closed-form expression, as derived in~\citesupp{app:mnltie}:
\begin{equation}
p(\mbf{x} \succ \mcl{C} \setminus \{\mbf{x}\}|  \mbf{f}_{\mcl{C} \cup \{\mbf{x}\}}; \delta) = \frac{e^{f_{\mbf{x}}}}{e^{f_{\mbf{x}}} + \sum_{\mbf{x}' \in \mcl{C} \setminus \{\mbf{x}\}} e^{f_{\mbf{x}'} + \delta}}
\label{eq:pref}
\end{equation}
where $p(\mbf{x} \succ \mcl{C} \setminus \{\mbf{x}\}) \triangleq p(\forall \mbf{x}' \in \mcl{C} \setminus \{\mbf{x}\}\quad\mbf{x} \succ \mbf{x}')$ and $\mbf{f}_{\mcl{C} \cup \{\mbf{x}\}} \triangleq (f_{\mbf{x}''})_{\mbf{x}'' \in \mcl{C} \cup \{\mbf{x}\}}$ consists of function values at inputs in $\mcl{C} \cup \{\mbf{x}\}$ \citep{cantillo2010thresholds}.
As a result, when $\delta = 0$, inputs with equal objective function values have equal probabilities of being preferred while inputs with higher objective function values are exponentially more likely to be preferred.
When there is no tie (i.e., $\delta = 0$), the model can be generalized to accept a ranking (i.e., $\mbf{x}_1 \succ \mbf{x}_2 \succ \dots \succ \mbf{x}_m$) as an observation following the Plackett-Luce model \citep{luce59,plackett1975analysis}:
\begin{equation}
\begin{array}{l}
\displaystyle p(\mbf{x}_1 \succ \mbf{x}_2 \succ \dots \succ \mbf{x}_m| \mbf{f}_{\cup_{i=1}^m \{\mbf{x}_i\}}) \vspace{1mm}\\
	 = \prod_{i=1}^{m-1} p(\mbf{x}_i \succ \cup_{j=i+1}^m \{\mbf{x}_j\}| \mbf{f}_{\cup_{j=i}^m \{\mbf{x}_j\}}; \delta=0)\ .
\end{array}
\label{eq:rank}
\end{equation}
In the following subsections, we introduce the notations for $2$ different types of observation: top-$k$ ranking and top-$1$ ranking with ties.

Apart from the difference in the noise model, the work of \citet{gonzalez2017preferential}, which defines the probability that an input $\mbf{x}$ is preferred over another input $\mbf{x}'$ as $p(\mbf{x} \succ \mbf{x}') = (1 + \exp(f_{\mbf{x}'} - f_{\mbf{x}}))^{-1}$, can be viewed as a special case of \eqref{eq:pref}. Therefore, by interpreting the formulation from the multinomial logit model with ties, our model is a generalization of the model in \cite{gonzalez2017preferential}.
\subsection{Top-$k$ Ranking}
\label{subsec:topk}
When there is no tie ($\delta = 0$), let a \emph{top-$k$ ranking} (i.e., a ranking of the top-$k$ inputs in a finite set) over a finite set $\mcl{C} \subset \mcl{X}$ (where $0 < k < |\mcl{C}|$) of inputs (i.e., interpreted as choices) be denoted as $\mbf{o}_{\mcl{C}}^k \triangleq \left\{\mbf{o}_{\mcl{C}}^k(i)\right\}_{i=1}^k$ which is an ordered set of $k$ inputs in descending order of preference in $\mcl{C}$, that is, $\mbf{o}_{\mcl{C}}^k(i) \in \mcl{C}$ and $\mbf{o}_{\mcl{C}}^k(i) \succ \mbf{o}_{\mcl{C}}^k(j)$ for all $i < j$. For example, $\mbf{o}^k_{\mcl{C}}(1)$ is the most preferred input in $\mcl{C}$, and similarly, $\mbf{o}^k_{\mcl{C}}(i)$ is the $i$-th most preferred input in $\mcl{C}$.
From \eqref{eq:rank}, the probability of a top-$k$ ranking is expressed as
\begin{equation}
p(\mbf{o}^k_{\mcl{C}}| \mbf{f}_{\mcl{C}}) = \prod_{i=1}^k p(\mbf{o}^k_{\mcl{C}}(i) \succ \mcl{C} \setminus \cup_{j=1}^i \{\mbf{o}^k_{\mcl{C}}(j)\}| \mbf{f}_{\mcl{C}}; \delta=0) \ .
\label{eq:topk}
\end{equation}
When $|\mcl{C}| = k + 1 = 2$, a top-$k$ ranking reduces to a pairwise preference (i.e., between a pair of inputs)  which is the observation considered in \cite{gonzalez2017preferential}.

Note that $\mbf{o}^{|\mcl{C}| - 1}_{\mcl{C}}$ is a (full) ranking of all inputs in~$\mcl{C}$.
Its probability (i.e., specified in \eqref{eq:topk} when $k = |\mcl{C}| - 1$) differs from that of a batch of pairwise preferences, the latter of which is the product of probabilities of pairwise preferences in the batch. This is different from the work of  \citet{gonzalez2017preferential} which claims that rankings can be trivially mapped to pairwise preferences. In fact, simple approaches of mapping rankings to pairwise preferences can violate a probability axiom, as shown in \citesupp{app:rankpairwise}.
\subsection{Top-$1$ Ranking with Ties}
\label{subsec:tie}
In practice, we are often incapable of stating a strict preference between inputs with similar utility function values. One may ignore the observation in this case if the model can only handle strict preference. However, overlooking this observation reduces the query efficiency of the BO algorithm.
Therefore, apart from the strict preference between inputs, tie/indifference should be allowed in the model to capture this observation.

In this subsection, we investigate one such possibility of ties (i.e., $\delta > 0$) in the observation when $k=1$ (i.e., top-$1$ ranking with ties).
To simplify notation, we denote $o_{\mcl{C}}$ as the only input in $\mbf{o}_{\mcl{C}}^1$, i.e., $o_{\mcl{C}}$ is the most preferred input in $\mcl{C}$.
Let $o_{\mcl{C}} = \varnothing$ denote the event where there exists a tie in finding the most preferred input in $\mcl{C}$. The probability of a tie can be expressed as follows:
\begin{equation}
\begin{array}{c}
p(o_{\mcl{C}} = \varnothing| \mbf{f}_{\mcl{C}}; \delta) = 1 - \sum_{o_{\mcl{C}} \in \mcl{C}} p(o_{\mcl{C}}| \mbf{f}_{\mcl{C}}; \delta)
\end{array}
\label{eq:tieprob}
\end{equation}
s.t.~$p\left(o_{\mcl{C}} | \mbf{f}_{\mcl{C}}; \delta\right) = p(o_{\mcl{C}} \succ \mcl{C} \setminus \{o_{\mcl{C}}\}| \mbf{f}_{\mcl{C}}; \delta)$ is specified in \eqref{eq:pref}. 
\begin{remark}
\emph{
When $k > 1$, the probability of a tie observation cannot be computed as straightforwardly as \eqref{eq:tieprob} because the tie relation between a pair of inputs (i.e., $\mbf{x} \sim \mbf{x}'$) is not transitive. Fig.~\ref{fig:nontransitive} shows a counter-example where $\mbf{x}_0 \sim \mbf{x}_1$ ($|u_{\mbf{x}_0} - u_{\mbf{x}_1}| < \delta$) and $\mbf{x}_1 \sim \mbf{x}_2$ ($|u_{\mbf{x}_1} - u_{\mbf{x}_2}| < \delta$) do not lead to $\mbf{x}_0 \sim \mbf{x}_2$ ($|u_{\mbf{x}_0} - u_{\mbf{x}_2}| < \delta$).
This is further elaborated in \citesupp{app:tieklarge1} and therefore left for future work.
}
\end{remark}
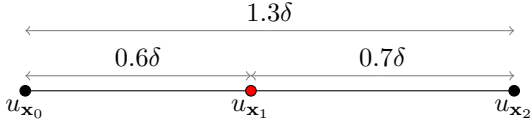
\begin{figure}
\centering
    \begin{tikzpicture}
    \draw[gray,<-] (0,0.2) -- (1.5,0.2); 
    \draw[black] (1.5,0.2) node[above,align=center]{$0.6\delta$};
    \draw[gray,->] (1.5,0.2) -- (3,0.2);
    \draw[gray,<-] (3,0.2) -- (4.75,0.2); 
    \draw[black] (4.75,0.2) node[above,align=center]{$0.7\delta$};
    \draw[gray,->] (4.75,0.2) -- (6.5,0.2);
    \draw[gray,<-] (0,0.8) -- (3.25,0.8); 
    \draw[black] (3.25,0.8) node[above,align=center]{$1.3\delta$};
    \draw[gray,->] (3.25,0.8) -- (6.5,0.8);

    \draw (0,0) node[align=center,   below] {$u_{\mbf{x}_0}$} -- (3,0) node[align=center, below] {$u_{\mbf{x}_1}$} -- (6.5,0) node[align=center,  below] {$u_{\mbf{x}_2}$};  
    \draw[black,fill=black] (0,0) circle (2pt);
    \draw[black,fill=red] (3,0) circle (2pt);
    \draw[black,fill=black] (6.5,0) circle (2pt);
    \end{tikzpicture}
    \caption{A counter-example based on the transitivity property of a tie relation between a pair of inputs.}
    \label{fig:nontransitive}
\end{figure}
\section{Variational Inference with\\ Top-$k$ Ranking Observations}
\label{sec:vi}
We employ a noiseless \emph{Gaussian process} (GP) to model the unknown objective function $f$. In other words, function values at every finite subset of $\mcl{X}$ follow a multivariate Gaussian distribution \cite{rasmussen06} whose \emph{prior} distribution is specified by the GP prior mean and covariance $k_{\mbf{x}, \mbf{x}'} \triangleq \text{cov}[f_{\mbf{x}}, f_{\mbf{x}'}]$. As the preference probability in \eqref{eq:pref} does not change when $f$ is shifted by a constant (i.e., shift-invariant), the GP prior mean is set to $0$. The covariance is defined by the \emph{squared exponential} kernel, i.e., $k_{\mbf{x}\mbf{x}'} \triangleq \sigma_s^2 \exp(-0.5 (\mbf{x} - \mbf{x}')^\top \Lambda^{-2} (\mbf{x} - \mbf{x}'))$ where the hyperparameters consist of the length-scales $\Lambda \triangleq \text{diag}[l_1, \dots, l_d]$ and the signal variance $\sigma_s^2$.

Unlike the work of \citet{gonzalez2017preferential} that models the function $f_{\mbf{x}'} - f_{\mbf{x}}$ (i.e., a function with the input as a pair $[\mbf{x}, \mbf{x}']$), we directly model the objective function $f_{\mbf{x}}$ which requires only half the dimension of the input of $f_{\mbf{x}'} - f_{\mbf{x}}$. 
Therefore, while our GP can be intuitively interpreted as the belief over the unknown objective function, the GP in \cite{gonzalez2017preferential} cannot.
As a result, in our model, the maximizer of the mean function of the posterior GP belief can be viewed as an estimate of the maximizer of the objective function, while the work of \citet{gonzalez2017preferential} needs to introduce the soft-Copeland score to estimate the maximizer of the objective function. Evaluating the soft-Copeland score requires the use of Monte-Carlo integration over $\mbf{x} \in \mcl{X}$, which is prohibitively expensive for problems with high input dimension.
Another advantage of our model is the ability of handling ranking observation, which is difficult to extend from the GP model of pairwise preferences in \cite{gonzalez2017preferential}, as explained in \citesupp{app:rankpairwise}.

Let $\mcl{D}$ denote the observations (e.g., top-$k$ rankings and top-$1$ rankings with ties) in a BO iteration, $\mcl{X}_{\mcl{D}}$ denote the set of distinct inputs in $\mcl{D}$, and $\mbf{f}_{\mcl{X}_\mcl{D}}$ denote the function values evaluated at $\mcl{X}_{\mcl{D}}$.
The likelihood $p(\mcl{D}|\mbf{f}_{\mcl{X}_\mcl{D}})$ of the observations is specified in Sec.~\ref{sec:mlm} and is not a Gaussian distribution. Hence, the GP posterior belief given the observations does not have a closed-form expression, unlike that of GP regression in conventional BO. 
To estimate the posterior belief, we use the variational inference technique to learn an approximate Gaussian posterior belief $q(\mbf{f}_{\mcl{X}_\mcl{D}}) \triangleq \mcl{N}(\mbf{f}_{\mcl{X}_\mcl{D}}| \bm{\mu}_{\mcl{X}_\mcl{D}}, \bm{\Sigma}_{\mcl{X}_\mcl{D}})$ of $\mbf{f}_{\mcl{X}_\mcl{D}}$ given $\mcl{D}$ by maximizing the \emph{evidence lower bound} (ELBO):\footnote{Like the definition of $\mbf{f}_{\mcl{X}_\mcl{D}}$, $\bm{\mu}_{\mcl{X}_\mcl{D}}$ is a vector of posterior mean values at $\mcl{X}_\mcl{D}$ and $\bm{\Sigma}_{\mcl{X}_\mcl{D}}$ is a covariance matrix whose elements are the posterior covariance between function values at $\mcl{X}_\mcl{D}$.}
\begin{equation}
\int\hspace{-0.5mm} q(\mbf{f}_{\mcl{X}_\mcl{D}}) \log p(\mcl{D}|\mbf{f}_{\mcl{X}_\mcl{D}})\ \text{d}\mbf{f}_{\mcl{X}_\mcl{D}}
    \hspace{-0.5mm}- \hspace{-0.5mm}\int\hspace{-0.5mm} q(\mbf{f}_{\mcl{X}_\mcl{D}}) \log \frac{q(\mbf{f}_{\mcl{X}_\mcl{D}})}{p(\mbf{f}_{\mcl{X}_\mcl{D}})}\ \text{d}\mbf{f}_{\mcl{X}_\mcl{D}} .
\label{eq:stoelbo}
\end{equation}
The ELBO can be maximized with a stochastic gradient ascent algorithm to obtain $\bm{\mu}_{\mcl{X}_\mcl{D}}$, $\bm{\Sigma}_{\mcl{X}_\mcl{D}}$, the GP hyperparameters, and the threshold $\delta$ if ties exist, i.e., by drawing a random mini-batch of $\mbf{f}_{\mcl{X}_\mcl{D}}$ from the current variational distribution $q(\mbf{f}_{\mcl{X}_\mcl{D}})$ in each iteration to optimize \eqref{eq:stoelbo}.\footnote{To ensure $\bm{\Sigma}_{\mcl{X}_\mcl{D}}$ is a positive-definite matrix, we optimize its square root lower-triangular matrix. A regularizer over the length-scales can be applied to ensure sufficiently large length-scales.}
Since our GP directly models the objective function, the length-scales of the GP can be interpreted as the rate of decay of the spatial correlation of the objective function in terms of the Euclidean distance between the inputs. 
Given that the approximate posterior belief $q(\mbf{f}_{\mcl{X}_\mcl{D}})$ is a multivariate Gaussian distribution $\mcl{N}(\bm{\mu}_{\mcl{X}_\mcl{D}}, \bm{\Sigma}_{\mcl{X}_\mcl{D}})$, the posterior predictive belief of the function values at any finite subset $\mcl{X}' \subset \mcl{X}$ given $\mcl{D}$ (i.e., by marginalizing out $\mbf{f}_{\mcl{X}_\mcl{D}}$) is also a multivariate Gaussian distribution whose mean and covariance matrix are specified as follows:
\begin{equation}
\begin{array}{r@{}l}
\bm{\mu}_{\mcl{X}'} \triangleq&\displaystyle\ \mbf{K}_{\mcl{X}'\mcl{X}_\mcl{D}} \mbf{K}_{\mcl{X}_\mcl{D} \mcl{X}_\mcl{D}}^{-1} \bm{\mu}_{\mcl{X}_\mcl{D}}\vspace{1mm}\\
\bm{\Sigma}_{\mcl{X}'} \triangleq&\displaystyle\ \mbf{K}_{\mcl{X}'\mcl{X}'}
	- \mbf{K}_{\mcl{X}' \mcl{X}_\mcl{D}} \bm{\Lambda} \mbf{K}_{\mcl{X}_\mcl{D} \mcl{X}'}
\end{array}
\label{eq:post}
\end{equation}
where $\mbf{K}_{\mcl{X}_\mcl{D} \mcl{X}_\mcl{D}} \triangleq (k_{\mbf{x}\mbf{x}'})_{\mbf{x},\mbf{x}' \in \mcl{X}_\mcl{D}}$, $\mbf{K}_{\mcl{X}'\mcl{X}'} \triangleq (k_{\mbf{x}\mbf{x}'})_{\mbf{x},\mbf{x}'\in \mcl{X}'}$, $\mbf{K}_{\mcl{X}' \mcl{X}_\mcl{D}} \triangleq (k_{\mbf{x}\mbf{x}'})_{\mbf{x} \in \mcl{X}', \mbf{x}' \in \mcl{X}_\mcl{D}}$, $\mbf{K}_{\mcl{X}_\mcl{D}\mcl{X}'} \triangleq \mbf{K}_{\mcl{X}' \mcl{X}_\mcl{D}}^\top$, and $\bm{\Lambda} \triangleq \mbf{K}_{\mcl{X}_\mcl{D} \mcl{X}_\mcl{D}}^{-1} (\mbf{K}_{\mcl{X}_\mcl{D} \mcl{X}_\mcl{D}} - \bm{\Sigma}_{\mcl{X}_\mcl{D}}) \mbf{K}_{\mcl{X}_\mcl{D} \mcl{X}_\mcl{D}}^{-1}$. Note that sparse GP models can be used to reduce the time complexity of $\mcl{O}(|\mcl{X}_{\mcl{D}}|^3)$ (i.e., arising from the matrix inversion $\mbf{K}_{\mcl{X}_\mcl{D} \mcl{X}_\mcl{D}}^{-1}$) \cite{quinonero2005unifying}, but the full GP model is used in this paper to be comparable with the models used by other baseline methods. 

\begin{figure}
\centering
\begin{tabular}{@{}cc@{}}
\begin{tabular}{@{}c@{}}
	\includegraphics[trim={8mm 3mm 4mm 0mm}, clip, height=0.15\textwidth]{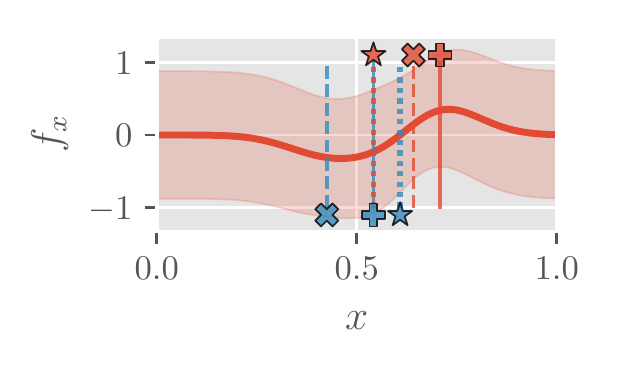}
	\\
	(a) GP posterior belief
	\\
	\includegraphics[trim={3mm 3mm 3mm 0.9mm}, clip, height=0.15\textwidth]{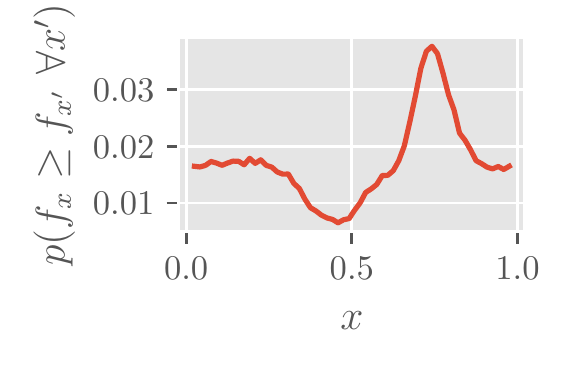}
	\\
	(b) Probability of maximizer
\end{tabular}
&\hspace{-5mm}
\begin{tabular}{@{}c@{}}
\begin{tikzpicture}
\node[anchor=south west,inner sep=0] at (0,0) {\includegraphics[trim={3mm 3mm 0mm 0mm}, clip, width=0.22\textwidth]{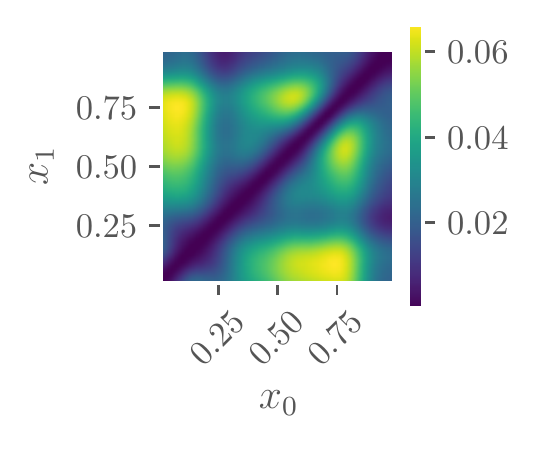}};
\draw[labelcolor,thick,<-] (2.1,2.55) -- (2.1,3.2) node[right,align=center]
    {Exploitation};
\draw[labelcolor,thick,<-] (1.15,2.4) -- (1.15,3.9) node[right,align=left]
    {Exploration};
\end{tikzpicture}
\\
(c) MPES of pairwise \\queries $\{x_0, x_1\}$
\end{tabular}
\end{tabular}
\caption{An example of top-$k$ BO with $3$ pairwise preferences: Inputs of each pair are represented as the $x$ values of cross, star, and plus markers. The red markers are the preferred inputs (i.e., plotted higher).}
\label{fig:example}
\end{figure}

Fig.~\ref{fig:example}a shows the GP posterior belief given the observations consisting of $3$ pairwise preferences. Due to noise, there is an incorrect preference between inputs plotted as stars, i.e., the input with the smaller $f_{\mbf{x}}$ is observed as being preferred. 
It can be observed that the difference in the function values at the inputs in the incorrect preference (plotted as stars) is smaller than that at the inputs in the correct preferences (crosses and pluses), which aligns with our formulation in~\eqref{eq:pref}.
Given the GP posterior belief, one can use the maximizer of the GP posterior mean function as an estimate of the maximizer of the objective function. In the next section, we utilize this GP posterior belief to design an information-theoretic acquisition function that can efficiently guide the query selection to search for the maximizer of the objective function.
\section{Multinomial Predictive Entropy Search (MPES)}
\label{sec:mpes}
While information-theoretic acquisition functions have been explored extensively in  conventional BO \citep{es,pes,ru2018fast,ppes,wang17mes}, they have not been investigated for preferential BO or our generalized top-$k$ BO.
Therefore, we propose to construct a principled acquisition function based on information theory to select the next query (i.e., a set $\mcl{C}$ of inputs), which we call \emph{multinomial predictive entropy search} (MPES).
Specifically, the next query is selected such that it maximizes the information gain on the maximizer $\mbf{x}_*$ of the objective function through observing the (top-$k$ ranking) observation at the query. Let $\mathbf{o}_{\mcl{C}}^k$  denote the observation given a query $\mcl{C}$. The observation can be either a top-$k$ ranking or a top-$1$ ranking with ties (Sec.~\ref{sec:mlm}). The information gain is measured by the mutual information between~$\mbf{x}_*$ and~$\mathbf{o}_{\mcl{C}}^k$, which is interpreted as the reduction in the entropy (uncertainty) of the maximizer $\mbf{x}_*$ given the observation $\mbf{o}_{\mcl{C}}^k$:
\begin{equation*}
I(\mbf{o}_{\mcl{C}}^k; \mbf{x}_*|\mcl{D}) \displaystyle\triangleq H(p(\mbf{x}_*| \mcl{D})) - \mbb{E}_{p(\mbf{o}_{\mcl{C}}^k|\mcl{D})} [H(p(\mbf{x}_*|\mcl{D},\mbf{o}_{\mcl{C}}^k)) ]
\end{equation*}
where $H(p(\mbf{x}_*| \mcl{D})) \triangleq -\int_{\mbf{x}_*} p(\mbf{x}_*| \mcl{D}) \log p(\mbf{x}_*| \mcl{D})\ \text{d}\mbf{x}_*$ denotes the entropy of $\mbf{x}_*$ and $\mbb{E}_{p(\mbf{o}_{\mcl{C}}^k|\mcl{D})}[H(p(\mbf{x}_*|\mcl{D},\mbf{o}_{\mcl{C}}^k))]$ denotes the conditional entropy of $\mbf{x}_*$ given $\mbf{o}_{\mcl{C}}^k$. However, the above expression requires a prohibitively expensive evaluation of $p(\mbf{x}_*|\mcl{D},\mbf{o}_{\mcl{C}}^k)$ for all possible values of $\mbf{o}_{\mcl{C}}^k$. This issue also exists in several information-theoretic acquisition functions in  conventional BO such as in  \citep{es,villemonteix2008informational}. Therefore, we employ the symmetric property of mutual information to express the acquisition function as
\begin{equation}
\hspace{-1.7mm}
\displaystyle\begin{array}{l}
\displaystyle I(\mbf{o}_{\mcl{C}}^k; \mbf{x}_*|\mcl{D}) 
= H(p(\mbf{o}_{\mcl{C}}^k| \mcl{D})) - \mbb{E}_{p(\mbf{x}_*|\mcl{D})} [H(p(\mbf{o}_{\mcl{C}}^k|\mcl{D},\mbf{x}_*))]\vspace{1mm}\\
=\displaystyle\ \sum_{\mbf{o}_{\mcl{C}}^k} \int_{\mbf{x}_*} p(\mbf{o}_{\mcl{C}}^k, \mbf{x}_*| \mcl{D}) \log \frac{p(\mbf{o}_{\mcl{C}}^k|\mcl{D}, \mbf{x}_*)}{p(\mbf{o}_{\mcl{C}}^k |\mcl{D})}\ \text{d}\mbf{x}_*\ .
\end{array}
\label{eq:mpes}
\end{equation}
The next query is selected as $\argmax_{\mcl{C} \in \mcl{X}^{|\mcl{C}|}} I(\mbf{o}_{\mcl{C}}^k; \mbf{x}_*|\mcl{D})$, which trades off between exploration (i.e., maximizing $H(p(\mbf{o}_{\mcl{C}}^k| \mcl{D}))$) and exploitation (i.e., minimizing $\mbb{E}_{p(\mbf{x}_*|\mcl{D})} [H(p(\mbf{o}_{\mcl{C}}^k|\mcl{D},\mbf{x}_*))]$).
To evaluate \eqref{eq:mpes}, we approximate the integration over the maximizer $\mbf{x}_* \in \mcl{X}$ as a summation over a finite set $\mcl{X}_* \subset \mcl{X}$ of possible maximizers. If $\mcl{X}$ is discrete and $|\mcl{X}|$ is sufficiently small, we can set $\mcl{X}_* = \mcl{X}$. On the other hand, we can construct $\mcl{X}_*$ by optimizing function samples drawn from the GP posterior belief given $\mcl{D}$  \cite{pes,wang17mes,rahimi08random}. 
For high dimensional problems, additive GP can be employed, as explained in \cite{wang17mes}.
Given the finite set $\mcl{X}_*$ of possible maximizers, \eqref{eq:mpes}~can be expressed as follows:
\begin{equation}
\begin{array}{l}
I(\mbf{o}_{\mcl{C}}^k; \mbf{x}_*|\mcl{D}) \\
	\displaystyle\approx \sum_{\mbf{x}_* \in \mcl{X}_*} p(\mbf{x}_*| \mcl{D}) \sum_{\mbf{o}_{\mcl{C}}^k} p(\mbf{o}^k_{\mcl{C}}| \mcl{D}, \mbf{x}_*) \log \frac{p(\mbf{o}^k_{\mcl{C}}| \mcl{D}, \mbf{x}_*)}{p(\mbf{o}^k_{\mcl{C}}| \mcl{D})}
\end{array}
\label{eq:mpescompute}
\end{equation}
where the probabilities are estimated with sampling in the following procedure:
\begin{enumerate}
\item Draw $n$ samples $\mbf{f}_{\mcl{C} \cup \mcl{X}_*} \sim p(\mbf{f}_{\mcl{C} \cup \mcl{X}_*}|\mcl{D})$ of function values at $\mcl{X}_*$ and $\mcl{C}$ where $p(\mbf{f}_{\mcl{C} \cup \mcl{X}_*}|\mcl{D})$ is the density of the multivariate Gaussian distribution specified in~\eqref{eq:post}.
\item Estimate the posterior probability of $\mbf{x}_*$ given $\mcl{D}$ as $p(\mbf{x}_*|\mcl{D}) = n^{-1} \sum_{\mbf{f}_{\mcl{X}_*}} \mbb{I}_{\mbf{x}_* = \argmax \mbf{f}_{\mcl{X}_*}}$ where the summation is taken over function samples at $\mcl{X}_*$ in step~$1$.
\item Estimate the joint posterior probability of $\mbf{o}^k_{\mcl{C}}$ and $\mbf{x}_*$ given $\mcl{D}$ as $p(\mbf{o}_{\mcl{C}}^k, \mbf{x}_*| \mcl{D}) = n^{-1} \sum_{\mbf{f}_{\mcl{C} \cup \mcl{X}_*}} \mbb{I}_{\mbf{x}_* = \argmax \mbf{f}_{\mcl{X}_*}} p(\mbf{o}_{\mcl{C}}^k| \mbf{f}_{\mcl{C}})$ where the summation is taken over function samples at $\mcl{C} \cup \mcl{X}_*$ obtained in step $1$ and the likelihood $p(\mbf{o}_{\mcl{C}}^k| \mbf{f}_{\mcl{C}})$ is described in Sec.~\ref{sec:mlm}.
\item Estimate the posterior probability of the observation $\mbf{o}^k_{\mcl{C}}$ given $\mcl{D}$ as $p(\mbf{o}_{\mcl{C}}^k|\mcl{D}) = \sum_{\mbf{x}_* \in \mcl{X}_*} p(\mbf{o}_{\mcl{C}}^k, \mbf{x}_*| \mcl{D})$ where $p(\mbf{o}_{\mcl{C}}^k, \mbf{x}_*| \mcl{D})$ is obtained in step $3$.
\item Estimate the posterior probability of the observation $\mbf{o}_{\mcl{C}}^k$ given $\mcl{D}$ and the maximizer $\mbf{x}_*$ as $p(\mbf{o}_{\mcl{C}}^k|\mcl{D}, \mbf{x}_*) = p(\mbf{o}_{\mcl{C}}^k, \mbf{x}_*| \mcl{D}) / p(\mbf{x}_*|\mcl{D})$ where $p(\mbf{o}_{\mcl{C}}^k, \mbf{x}_*| \mcl{D})$ and $p(\mbf{x}_*|\mcl{D})$ are obtained in steps $3$ and $2$, respectively.
\end{enumerate}
Note that the number of possible top-$1$ rankings~$\mbf{o}_\mcl{C}^1$ only grows linearly w.r.t.~$|\mcl{C}|$. So, the evaluation of MPES can scale well to large $|\mcl{C}|$ for $k = 1$. When $k > 1$, the number of possible~$\mbf{o}_\mcl{C}^k$ grows exponentially w.r.t.~$|\mcl{C}|$. So, the cost of evaluating MPES is dominated by the sum over $|\mcl{C}|!/(|\mcl{C}| - k)!$ possible observations.
Therefore, we mainly focus on a small $|\mcl{C}|$ or $k = 1$ such as $|\mcl{C}| = 4$ in our experiments where we enumerate all possible $\mbf{o}_{\mcl{C}}^k$ to compute MPES.
In this case, we search for $\argmax_{\mcl{C} \in \mcl{X}^{|\mcl{C}|}} I(\mbf{o}_{\mcl{C}}^k; \mbf{x}_*|\mcl{D})$ by randomly selecting a number of subsets of $|\mcl{C}|$ inputs in $\mcl{X}$, which is empirically shown to significantly outperform EI and DTS in our experiments (Sec.~\ref{sec:experiment}).
Alternatively, DIRECT \cite{jones1993lipschitzian} may be used to optimize MPES. 
Nonetheless, searching over a high dimensional space is a challenging problem.
A potential direction to evaluate MPES for large $|\mcl{C}|$ and $k > 1$ is to express \eqref{eq:mpescompute} as
\begin{equation*}
\begin{array}{l}
I(\mbf{o}_{\mcl{C}}^k; \mbf{x}_*|\mcl{D}) \vspace{1mm}\\
	\displaystyle\approx\hspace{-0.5mm} \sum_{\mbf{x}_* \in \mcl{X}_*} p(\mbf{x}_*| \mcl{D}) \mbb{E}_{p(\mbf{o}^k_{\mcl{C}}| \mcl{D}, \mbf{x}_*)} \left[ \log p(\mbf{o}^k_{\mcl{C}}| \mcl{D}, \mbf{x}_*) / p(\mbf{o}^k_{\mcl{C}}| \mcl{D}) \right]
\end{array}
\end{equation*}
where the expectation 
$\mbb{E}_{p(\mbf{o}^k_{\mcl{C}}| \mcl{D}, \mbf{x}_*)}$
is approximated by stochastic sampling, i.e., drawing a number of $\mbf{o}^k_{\mcl{C}}$ following $p(\mbf{o}^k_{\mcl{C}}| \mcl{D}, \mbf{x}_*)$. It is performed by sampling $\mbf{o}^k_{\mcl{C}}$ from $p(\mbf{o}^k_{\mcl{C}}| \mcl{D}, \mbf{x}_*, \mbf{f}_{\mcl{C} \cup \mcl{X}_\star}) = p(\mbf{o}^k_{\mcl{C}}| \mbf{f}_{\mcl{C} \cup \mcl{X}_\star})$ (i.e., the likelihood function in Sec.~\ref{sec:mlm}) where $\mbf{f}_{\mcl{C} \cup \mcl{X}_\star}$ are samples in the above step~$1$ s.t.~$\mbf{x}_* = \argmax \mbf{f}_{\mcl{X}_*}$. The rationale is to estimate the expectation with samples $\mbf{o}_{\mcl{C}}^k$ of high probabilities $p(\mbf{o}^k_{\mcl{C}}| \mcl{D}, \mbf{x}_*)$ instead of with all of the possible $\mbf{o}_{\mcl{C}}^k$ for large $|\mcl{C}|$ and $k > 1$.

While joint optimization over the query requires searching over a large space (i.e., $\mcl{X}^{|\mcl{C}|}$) as compared to optimizing the inputs of a query one a time greedily (i.e., $\mcl{X}$) in \cite{brochu10tut,dewancker18,gonzalez2017preferential}, the latter ignores the correlation between function values evaluated at different inputs of the query when selecting the first input. This leads to inferior performance, as shown empirically in our experiments (Sec.~\ref{sec:experiment}). Since the primary goal of BO is to reduce the cost of obtaining the query observation, the BO performance should not be sacrificed for the cost of optimizing the acquisition function. This is the motivation for us to develop a BO algorithm that is capable of jointly optimizing over the query to improve the quality of the queries.
\begin{example}[Exploitation vs. exploration]
\emph{
Fig.~\ref{fig:example}c shows the MPES values for pairs of inputs in the domain $\mcl{X}$ given the GP posterior belief in Fig.~\ref{fig:example}a. The MPES values are symmetric about the line $x_0 = x_1$,  which is expected as the role of $x_0$ and $x_1$ are interchangeable. There are two regions with high MPES values, which are annotated as exploitation and exploration. The exploitation region is the region where the posterior means of both inputs in the pair are large. These inputs also have high probabilities of being the maximizer in Fig.~\ref{fig:example}b.
The exploration region is the region where the posterior mean of one input in the pair is large (i.e., its probability of being the maximizer is high in Fig.~\ref{fig:example}b) and the other input is far away from all inputs in the observed pairwise preferences (i.e., its probability of being the maximizer is not high in Fig.~\ref{fig:example}b). Thus, MPES is able to balance exploration and exploitation naturally without any explicit modification.
}
\end{example}
The joint optimization over the query and the exploration-exploitation trade-off of MPES contrast with existing approaches: \emph{expected improvement} (EI)~\cite{brochu10tut,dewancker18} and \emph{dueling-Thompson sampling} (DTS)~\cite{gonzalez2017preferential}.
In EI, the two inputs of the query (i.e., a pair of inputs) are selected independently: an input maximizing EI and the other input maximizing $\mu_{\mbf{x}}$.
Note that EI potentially leads to excessive exploitation \cite{gonzalez2017preferential}.
On the other hand, DTS selects inputs of a query one at a time: the first input maximizing the soft-Copeland score and the second input maximizing the variance of the preference given the first input. 
Furthermore, exploration is explicitly introduced by selecting the first input based on only $1$ sample from the GP belief using continuous Thompson sampling.
\section{Experiments and Discussion}
\label{sec:experiment}
In this section, we empirically demonstrate (a) the performance of our MPES in comparison with existing methods: \emph{expected improvement} (EI) \cite{mockus1978application} and \emph{dueling-Thompson sampling} (DTS) \cite{gonzalez2017preferential} using pairwise preferences in Sec.~\ref{subsec:bopairwise}, (b) the performance of MPES and the model of ties in Sec.~\ref{subsec:botie}, and (c) the performance of MPES with top-$k$ ranking in Sec.~\ref{subsec:botopk}.
For the experimental results of MPES, we label MPES with the values of $k$, $|\mcl{C}|$, and $\delta$: For example, MPES with $k=1$, $|\mcl{C}|=3$, and $\delta=0$ (i.e., top-$1$ ranking of a set of $3$ inputs without ties) is labeled as \emph{top-$1$ of $3$ $\delta=0$}. When $\delta > 0$ (i.e., ties exist), $\delta$ is unknown to our model and is obtained by optimizing~\eqref{eq:stoelbo}. To evaluate MPES, we set $|\mcl{X}_*|$ to $20$ and the number of samples is $n=1000$.
The code is available at \url{https://github.com/sebtsh/Top-k-Ranking-Bayesian-Optimization}.

Following the work of~\citet{pes}, we compute the \emph{immediate regret} in each BO iteration as the performance metric. For synthetic functions, it is the difference between the global maximum value $\max_{\mbf{x} \in \mcl{X}} f_{\mathbf{x}}$ of the objective function and 
the function value at an estimate of the maximizer from the GP belief.
This estimate of the maximizer is the maximizer of the GP posterior mean function (i.e., $\argmax_{\mbf{x} \in \mcl{X}} \mu_{\mbf{x}}$) for MPES and EI, while it is the maximizer of the soft-Copeland score for DTS.
A smaller immediate regret is preferred and indicates higher query efficiency. We repeat each experiment $10$ times to plot the average and  standard error of the immediate regret.

These acquisition functions are evaluated on $3$ synthetic benchmark functions with varying levels of difficulty and number of dimensions: (a) the simple 1-D Forrester function \cite{forrester2008engineering}, (b) the 2-D \emph{six-hump camel} (SHC) function ($6$ local minima) with the input domain restricted to $[-1.5,1.5]$ in each dimension \cite{molga2005test}, and (c) the $3$-D Hartmann function.\footnote{Available at \url{http://www-optima.amp.i.kyoto-u.ac.jp/member/student/hedar/Hedar_files/TestGO_files/Page1488.htm}.} These functions are originally modeled for finding the global minimum. However, since throughout this work, we have framed the problem as one of finding the global maximum, we take the negative values of these functions instead. 
The numbers of initial observations provided to the BO algorithms are $5$, $6$, and $12$ for experiments with the Forrester, SHC, and  Hartmann functions, respectively. We also perform experiments on the following $2$ real-world datasets:
\subsubsection{CIFAR-10 dataset} 
\begin{figure}
    \centering
    \includegraphics[trim={10mm 12mm 1.4cm 3mm}, clip, width=0.25\textwidth]{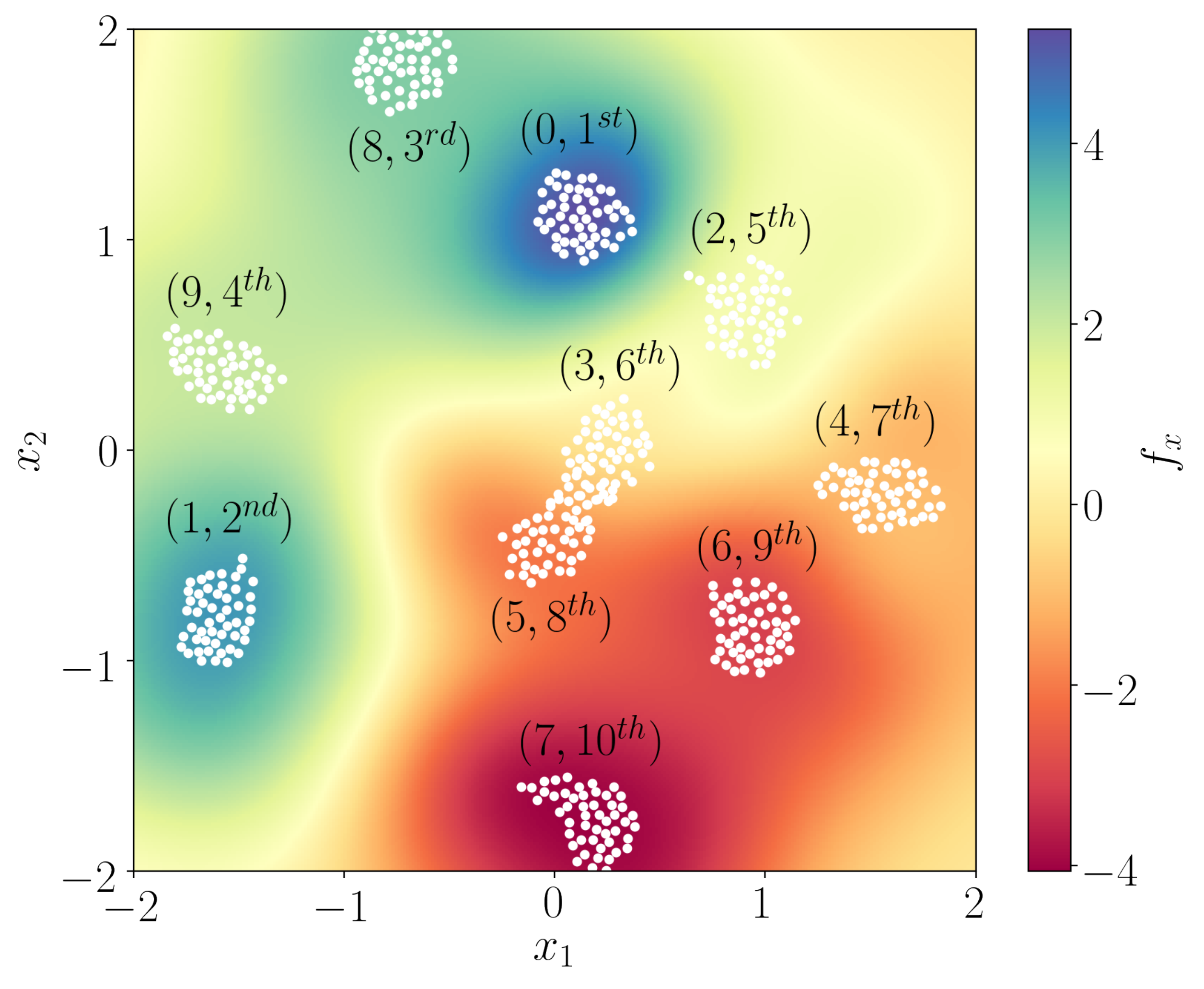}
    \caption{Plot of a smooth function of the ground truth ranking with the $2$-D embeddings of a subset of the CIFAR-$10$ dataset. Tuples in each cluster indicate the class number, followed by the order of the class in the ground truth ranking.}
    \label{fig:cifar10_pref}
\end{figure}

The input domain consists of $50000$ training images of the CIFAR-$10$ dataset~\cite{Krizhevsky09learningmultiple} which includes $32\times32$ colour images in $10$ classes. We use the following ground truth ranking of preference between classes: $7 \, \text{(horse)} \prec 6 \, \text{(frog)}  \prec 5 \, \text{(dog)} \prec 4 \, \text{(deer)} \prec 3 \, \text{(cat)}  \prec 2 \, \text{(bird)} \prec 9  \, \text{(truck)} \, \prec 8 \, \text{(ship)} \prec 1 \, \text{(automobile)} \prec 0 \, \text{(airplane)}$.
The objective is to identify the most preferred class through observing preferences/rankings between different images.
Given the GP posterior belief, the most preferred class is defined as the class where the average of the posterior mean of all images in the class is the maximum. 
The immediate regret is defined as the distance from the most preferred class given the GP posterior belief to class $0$ (i.e., airplane) in the ground truth ranking. For example, the immediate regret is $3$ if the most preferred class given the GP posterior belief is class $9$ (i.e., truck). So, the immediate regret is an integer in the range $[0, 9]$.
We reduce the dimensionality of CIFAR-$10$ dataset to an embedding space of $2$ dimensions with a combination of transfer learning from a CNN and a UMAP reduction \cite{mcinnes2018umap-software}. 
The embedding is visualized by plotting a smooth function of the ground truth ranking with the embeddings of images as the inputs in Fig.~\ref{fig:cifar10_pref}. 
We observe that this embedding separates the $10$ classes of CIFAR-$10$ into different clusters while preserving the relative distance between images such that visually similar images are close in the embedding, and vice versa.
For example, classes $3$ (i.e., cat) and $5$ (i.e., dog) are close to each other due to cats and dogs being visually similar. 
The objective function maps the $2$-D embedding of an image to the order of its class in the ground truth ranking subtracted by $5$ (e.g., the function value of a horse image is $1 - 5 = -4$).
Six initial observations are provided to the BO algorithms.%\vspace{0.4mm}
\subsubsection{SUSHI preference dataset}
Inspired from our dining example in Sec.~\ref{sec:intro}, this experiment is about learning the most preferred type of sushi through ranking observations. The objective function is generated from the real-world SUSHI preference dataset \cite{kamishima2003nantonac}, which is widely used for the evaluation of preference and ranking methods \cite{khetan2016data,vitelli2017probabilistic}. It consists of data for $100$ kinds of sushi and $5000$ user ratings of subsets of sushi. 
The input $\mbf{x}$ consists of $6$ features of the sushi. The objective function is obtained by scaling and shifting the average rating scores of users (which represents their average opinion) to the range $[-4,5]$ that is similar to the CIFAR-$10$ experiment.
The immediate regret is calculated as the distance between the BO algorithm's best guess of the most preferred sushi and the actual most preferred sushi in the ground truth ranking (based on the average rating scores). For example, if salmon sushi has rank $10$ in the ground truth ranking and the BO algorithm's best guess of the most preferred sushi at a timestep is salmon sushi, the algorithm's immediate regret at that timestep would be $9$ because salmon sushi is $9$ places away from the top. 
In these experiments, there are $10$ initial observations provided to the BO algorithms.
\subsection{BO with Pairwise Preferences}
\label{subsec:bopairwise}
\begin{figure}
\centering
\begin{tabular}{@{}c@{}c@{}}
\includegraphics[trim={0 4mm 0 2mm}, clip, height=0.146\textwidth]{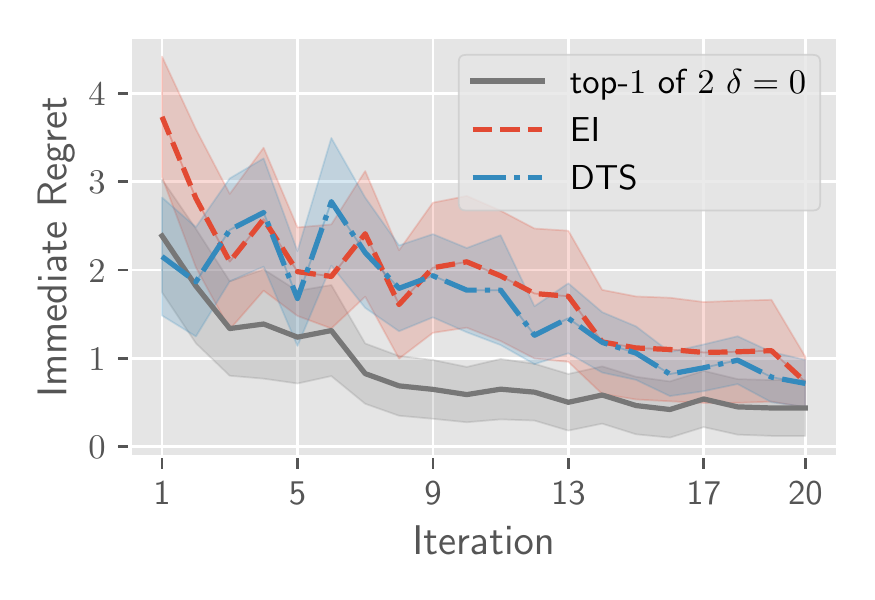}
&
\includegraphics[trim={0 4mm 0 2mm}, clip, height=0.146\textwidth]{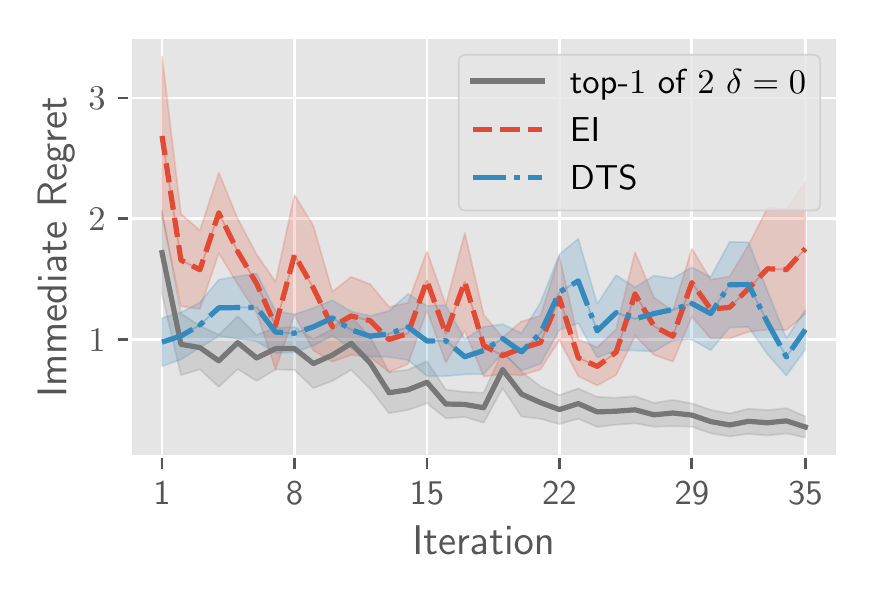}
\\
(a) Forrester function
&
(b) SHC function
\\
\includegraphics[trim={0 4mm 0 2mm}, clip, height=0.146\textwidth]{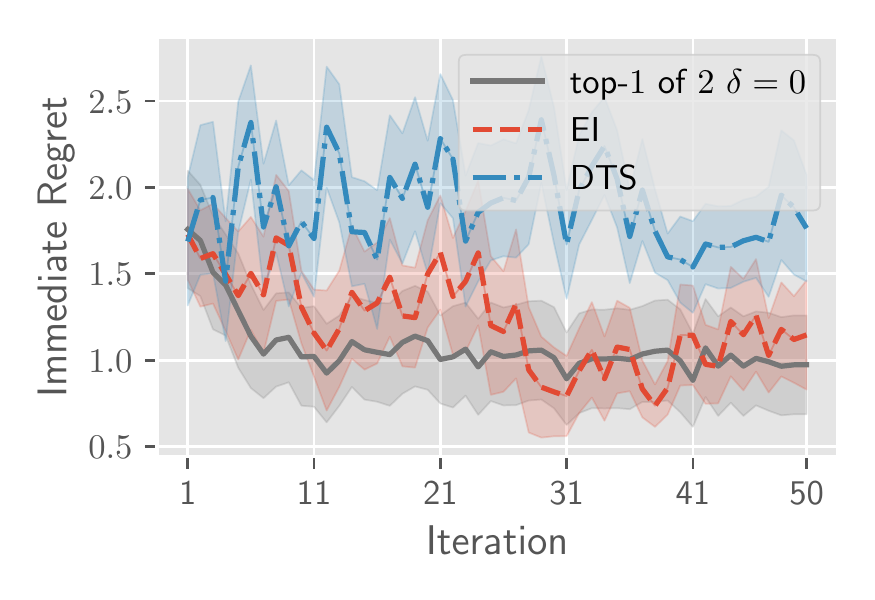}
&
\includegraphics[trim={0 4mm 0 2mm}, clip, height=0.146\textwidth]{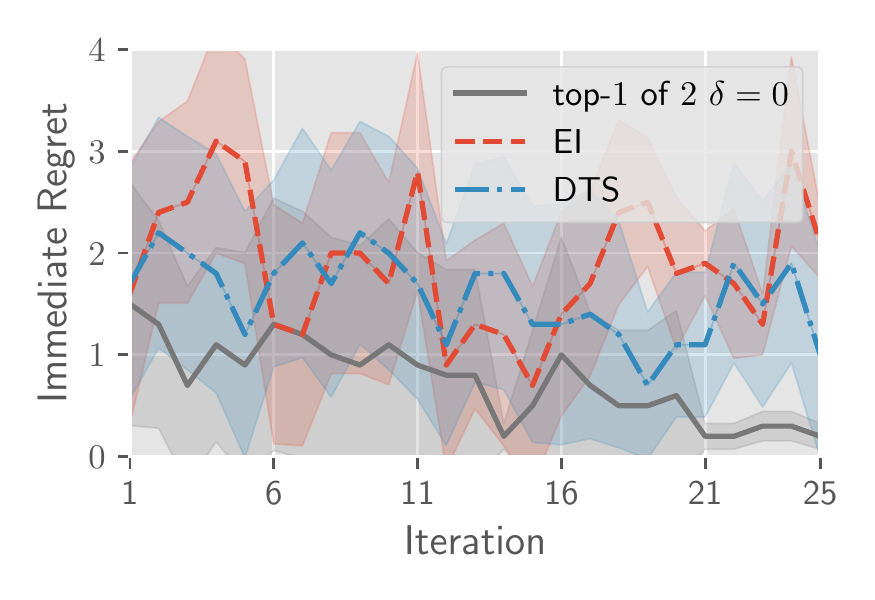}
\\
(c) Hartmann function
&
(d) CIFAR-$10$ dataset
\\
\includegraphics[trim={0 4mm 0 2mm}, clip, height=0.146\textwidth]{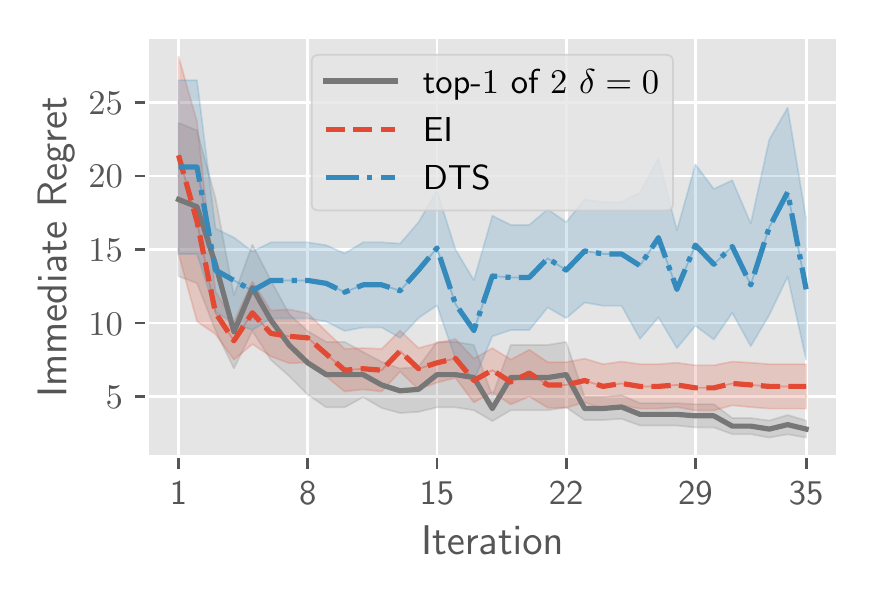}
&
\\
(e) SUSHI dataset
&
\end{tabular}
\caption{Plots of immediate regrets for experiments with pairwise preferences.}
\label{fig:bopairwise}
\end{figure}
Since the existing EI and DTS approaches are only able to handle pairwise preferences, we compare the performance of our MPES with EI and DTS using pairwise preferences here.
The immediate regrets for the Forrester, SHC, Hartmann, CIFAR-$10$, and SUSHI are shown in Fig.~\ref{fig:bopairwise}. It can be observed that MPES consistently outperforms both EI and DTS in these experiments. DTS outperforms EI in optimizing the Forrester function with $1$-D inputs (Fig.~\ref{fig:bopairwise}a). However, as the input dimension increases to $3$ in the optimization of the Hartmann function (Fig.~\ref{fig:bopairwise}c) and to $6$ in the optimization problem with the SUSHI dataset (Fig.~\ref{fig:bopairwise}e), DTS is outperformed by EI. This is due to the disadvantage of the model used by DTS whose input dimension is doubled with respect to the original dimension of the problem, as discussed in Sec.~\ref{sec:vi}.
\subsection{BO with Ties}
\label{subsec:botie}
In this subsection, we empirically show the competitive performance of MPES with tie observations by comparing with the existing DTS and EI using pairwise preferences.
On top of that, we let DTS and EI have an unfair advantage over MPES by having access to strict preferences (i.e., no tie) regardless of how close the utility values of the input pair are, while MPES can only receive a tie observation (i.e., there is no information about the preferred input in the pair) if the difference in the utility values of the input pair is less than $\delta$. 
Even so, MPES with the model of ties is able to outperform  DTS and EI both in optimizing synthetic benchmark functions and on the real-world datasets in Figs.~\ref{fig:boties}a-e.
This empirically shows the performance of both the model of ties in the likelihood (Sec.~\ref{subsec:tie}) and the performance of MPES.
\subsection{BO with Rankings}
\label{subsec:botopk}
\begin{figure}[t!]
\begin{tabular}{@{}c@{}c@{}}
\includegraphics[trim={0 4mm 0 2mm}, clip, height=0.16\textwidth]{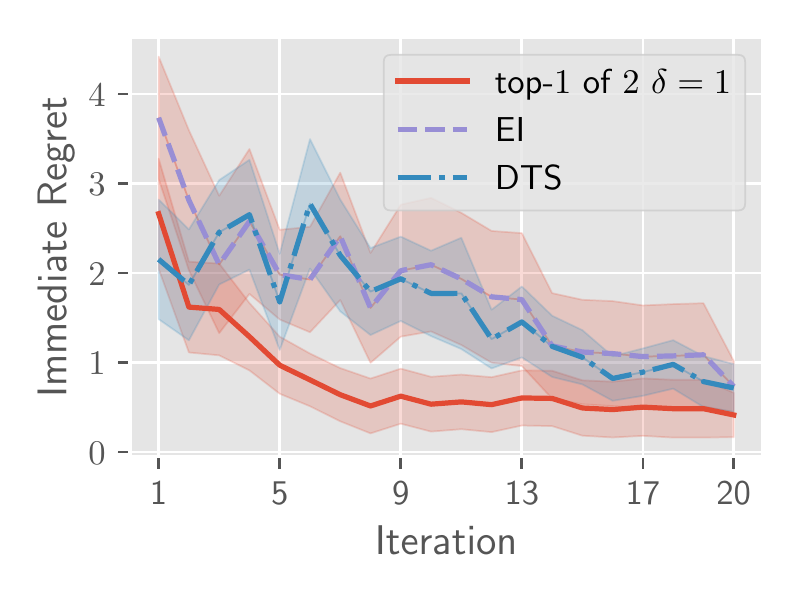}
&
\includegraphics[trim={0 4mm 0 2mm}, clip, height=0.16\textwidth]{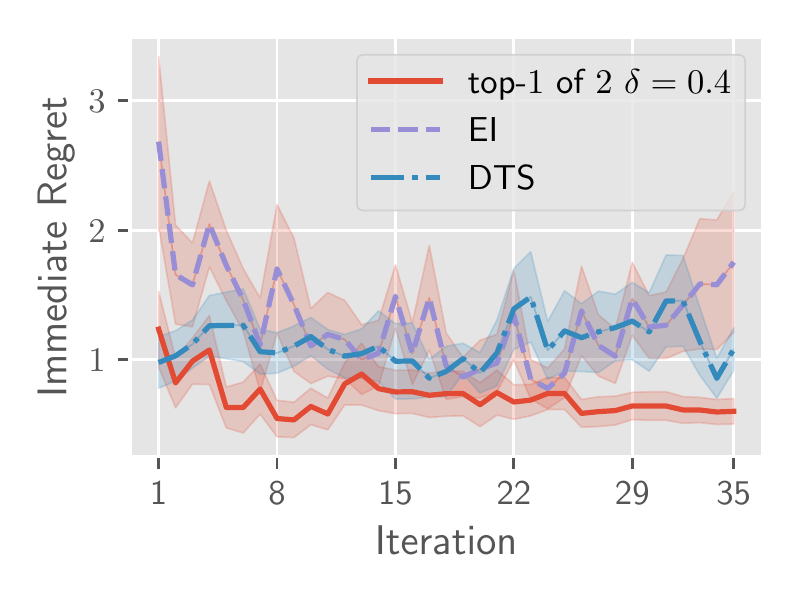}
\\
(a) Forrester function
&
(b) SHC function
\\
\includegraphics[trim={0 4mm 0 2mm}, clip, height=0.16\textwidth]{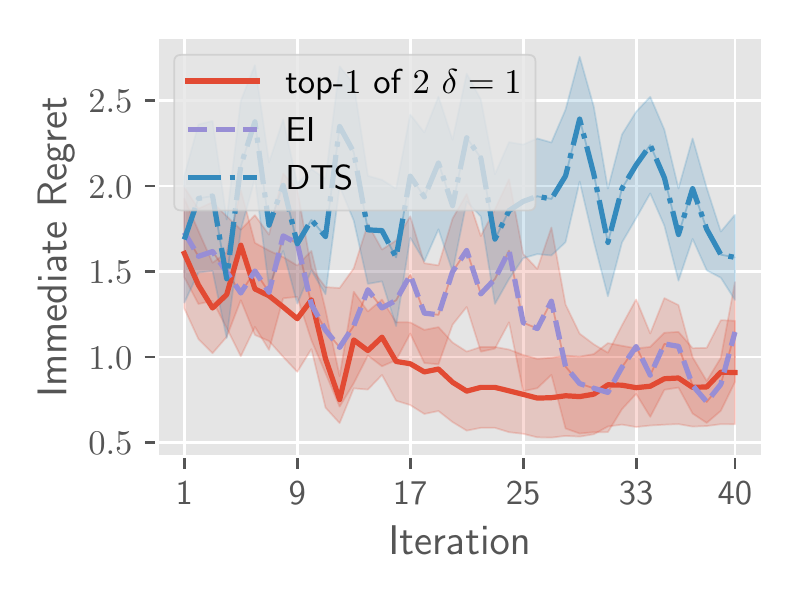}
&
\includegraphics[trim={0 4mm 0 2mm}, clip, height=0.16\textwidth]{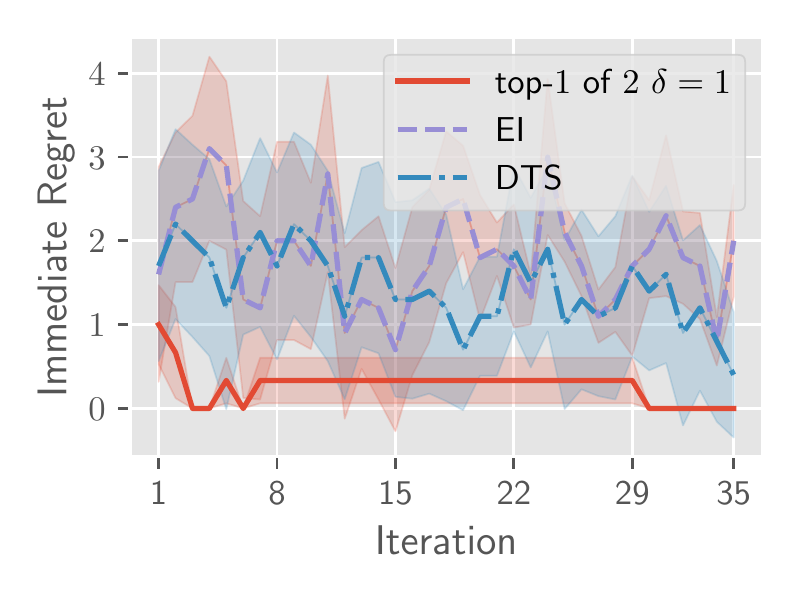}
\\
(c) Hartmann function
&
(d) CIFAR-$10$ dataset
\\
\includegraphics[trim={0 4mm 0 2mm}, clip, height=0.16\textwidth]{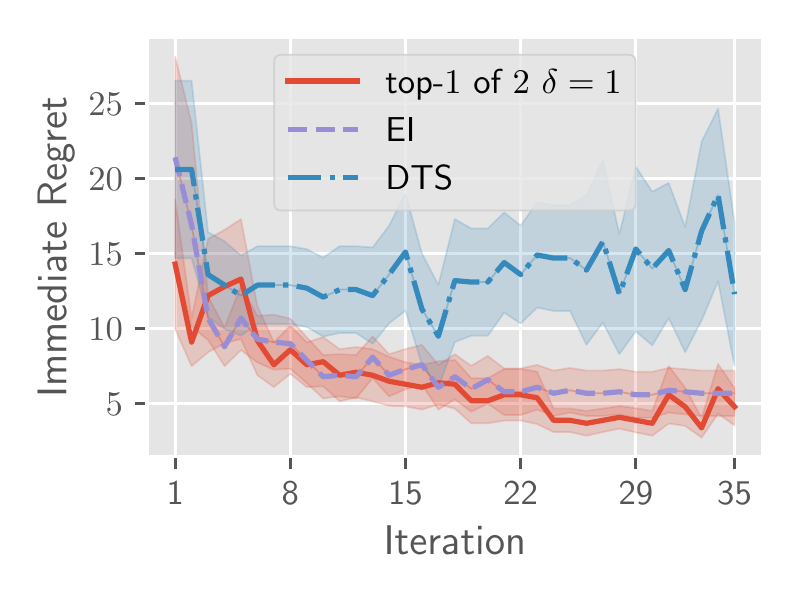}
&
\includegraphics[trim={0 4mm 0 2mm}, clip, height=0.16\textwidth]{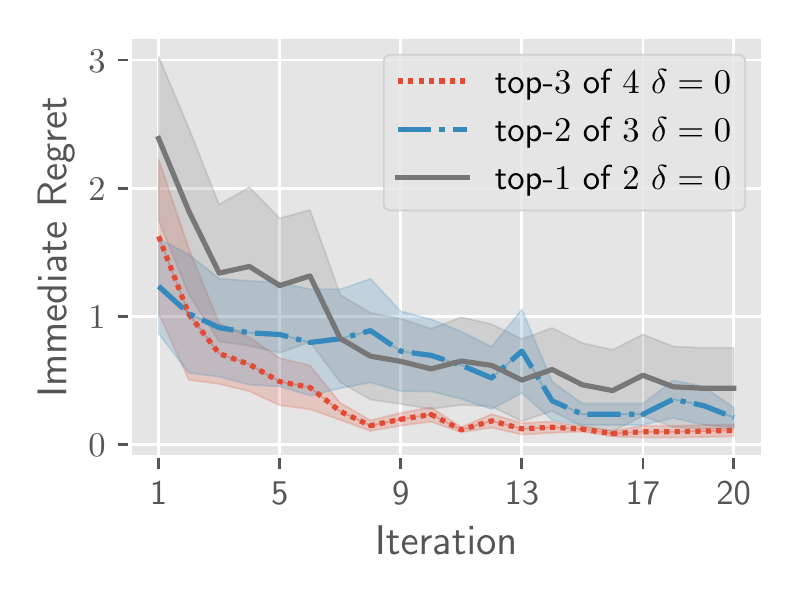}
\\
(e) SUSHI dataset
&
(f) Forrester function
\\
\includegraphics[trim={0 4mm 0 2mm}, clip, height=0.16\textwidth]{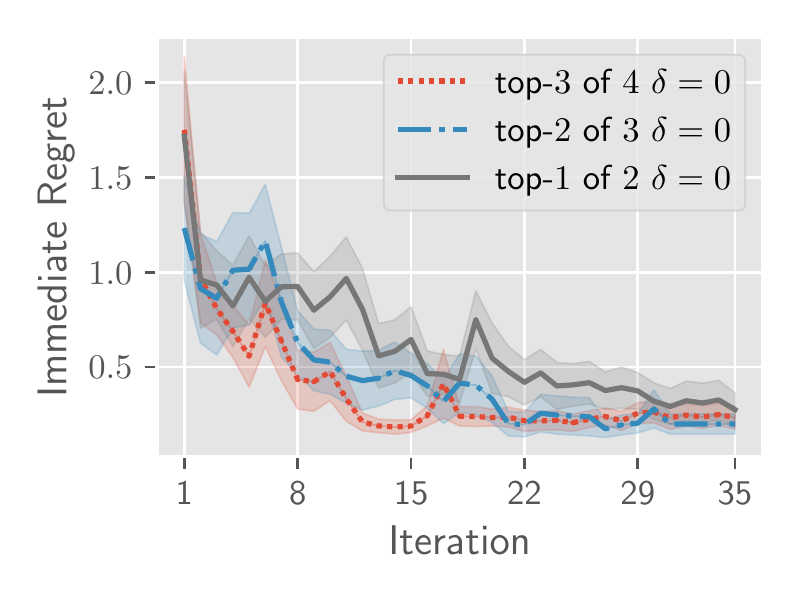}
&
\includegraphics[trim={0 4mm 0 2mm}, clip, height=0.16\textwidth]{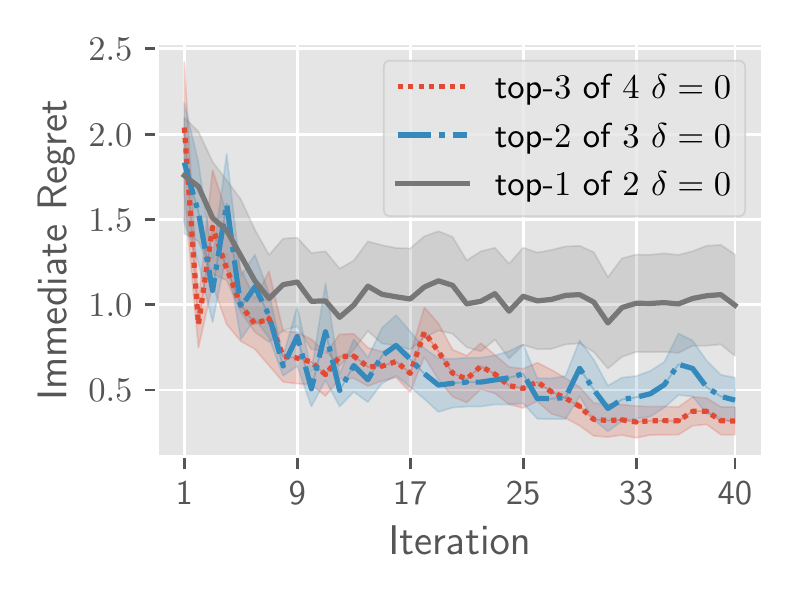}
\\
(g) SHC function
&
(h) Hartmann function
\\
\includegraphics[trim={0 4mm 0 2mm}, clip, height=0.16\textwidth]{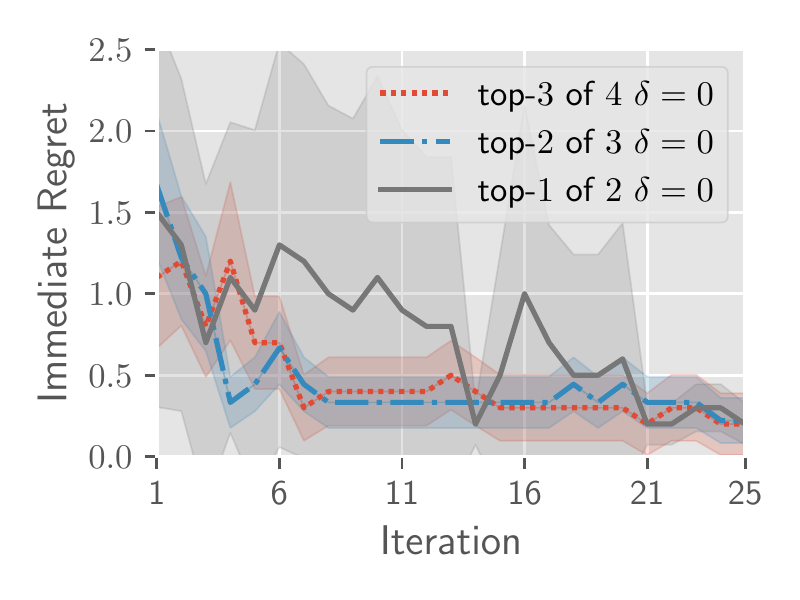}
&
\includegraphics[trim={0 4mm 0 2mm}, clip, height=0.16\textwidth]{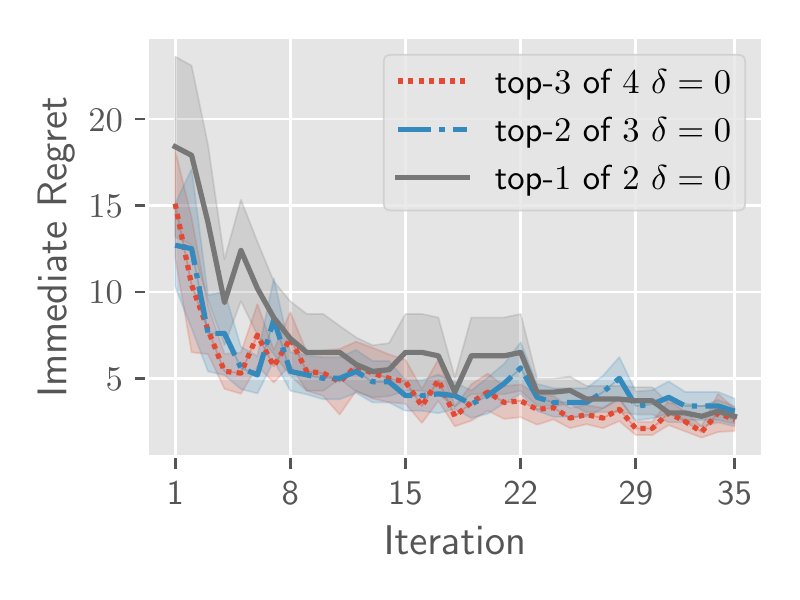}
\\
(i) CIFAR-$10$ dataset
&
(j) SUSHI dataset
\end{tabular}
\caption{Plots of immediate regrets for experiments with (a-e) pairwise preferences and (f-j) rankings. The observation of MPES can include ties while that of EI and DTS only include strict preferences (i.e., no tie). The threshold $\delta$ is learned from observation.}
\label{fig:boties}
\end{figure}

In this subsection, we empirically illustrate the advantage of ranking observations over pairwise preferences. In particular, as the rankings give more information about the GP posterior belief of the objective function, BO with ranking observation is expected to outperform BO with pairwise preferential observation given the same number of queries.
This is illustrated in Figs.~\ref{fig:boties}f-j where BO with $|\mcl{C}| > 2$ outperforms BO with $|\mcl{C}| = 2$ (i.e., pairwise preferences).
Furthermore, as $|\mcl{C}|$ increases, the performance of our algorithm improves.
\section{Conclusion}
\label{sec:conclusion}
This paper describes a principled approach to top-$k$ BO in both modeling the posterior belief of the objective function and formulating the acquisition function.
Inspired by the classic multinomial logit model, our model is capable of handling real-world observations including top-$k$ rankings of inputs and the existence of ties. 
Furthermore, based on an information-theoretic measure, we design the acquisition function of MPES that is capable of guiding the query selection through jointly optimizing all inputs of a query and balancing the exploration-exploitation trade-off.
Our new model and MPES are empirically demonstrated to have superior performance compared with existing approaches in several synthetic benchmark functions, CIFAR-$10$ dataset, and SUSHI preference dataset.
For future work, we plan to generalize MPES to nonmyopic BO~\citep{dmitrii20a,ling16}, 
batch BO~\citep{daxberger17}, high-dimensional BO~\citep{NghiaAAAI18}, private outsourced BO~\citep{dmitrii20b}, and multi-fidelity BO~\citep{yehong17,ZhangUAI19} settings and incorporating early stopping~\citep{dai2019} and recursive reasoning~\citep{dai2020}.
\section*{Broader Impact}
There are many real-world applications (e.g., food/movie/ music preference, art aesthetics, interior design, and place of interests (tourism)) where the objective function cannot be directly evaluated. In these applications, top-$k$ ranking BO is a promising optimization method (e.g., finding the best food recipe, the most pleasing design, and the most attractive place for tourists) with a limited budget of queries/trials.
Furthermore, the possibility of different types of observation in our models provides more options to design the data collection process, which potentially improves the user experience. 

There are also several considerations in applying our model. The negative impacts of our method can happen due to the poor performance when the underlying assumptions of our model (e.g., the Gumbel noise and the independence of irrelevant alternatives (inherent in the logit model)) are violated. Therefore, an application is required to examine if these assumptions are satisfied. In case they do not hold, further investigation into alternative models is necessary.
For applications with constraints (e.g., a combination of ingredients can cause food poisoning or be unsafe during pregnancy), we need to extend the current work to incorporate the constraints into the optimization process.

In brief, we believe our work is beneficial to the society when the above issues are taken into consideration. It is a step forward to extend the use of BO to daily applications.

\subsubsection{Acknowledgments.} This research/project is supported by A*STAR under its RIE$2020$ Advanced Manufacturing and Engineering (AME) Industry Alignment Fund -- Pre Positioning (IAF-PP) (Award A$19$E$4$a$0101$). 

\bibliography{bo}

\appendix 

\onecolumn
\section{Derivation of the Probability of Preference with Ties}
\label{app:mnltie}
In this section, we derive the probability of preference with ties \eqref{eq:pref}.
Recall that $u_{\mbf{x}} = f_{\mbf{x}} + \epsilon_{\mbf{x}}$ where $\epsilon_{\mbf{x}}$ follows a Gumbel distribution with parameters: $\mu = 0$ and $\beta = 1$. Hence, the noise p.d.f., denoted as $p(\epsilon_{\mbf{x}})$, and c.d.f., denoted as $F(\epsilon_{\mbf{x}})$, are
\begin{align*}
p(\epsilon_{\mbf{x}}) &= e^{-\epsilon_{\mbf{x}}} e^{-e^{-\epsilon_{\mbf{x}}}} \text{ and}\\
F(\epsilon_{\mbf{x}}) &= e^{-e^{-\epsilon_{\mbf{x}}}}\ ,
\end{align*}
respectively. Then, we can compute the probability that $\mbf{x}$ is preferred over $\mcl{C} \setminus \{\mbf{x}\}$ given $\mbf{f}_{\mcl{C} \cup \{\mbf{x}\}}$  as follows:
\begin{align*}
&p(\mbf{x} \succ \mcl{C} \setminus \{\mbf{x}\}| \mbf{f}_{\mcl{C} \cup \{\mbf{x}\}}; \delta)\\
    &= p(\forall\mbf{x}' \in \mcl{C}\setminus \{\mbf{x}\}\quad u_{\mbf{x}} \ge u_{\mbf{x}'} + \delta \mid \mbf{f}_{\mcl{C} \cup \{\mbf{x}\}})\\
    &= p(\forall\mbf{x}' \in \mcl{C}\setminus \{\mbf{x}\}\quad f_{\mbf{x}} + \epsilon_{\mbf{x}} \ge f_{\mbf{x}'} + \epsilon_{\mbf{x}'} + \delta\mid \mbf{f}_{\mcl{C} \cup \{\mbf{x}\}})\\
    &= p(\forall\mbf{x}' \in \mcl{C}\setminus \{\mbf{x}\}\quad \epsilon_{\mbf{x}'} \le \epsilon_{\mbf{x}} + f_{\mbf{x}} - f_{\mbf{x}'} - \delta \mid \mbf{f}_{\mcl{C} \cup \{\mbf{x}\}})\\
    &= \int_{-\infty}^{\infty}
        p(\epsilon_{\mbf{x}})\  p(\forall\mbf{x}' \in \mcl{C}\setminus \{\mbf{x}\}\quad \epsilon_{\mbf{x}'} \le \epsilon_{\mbf{x}} + f_{\mbf{x}} - f_{\mbf{x}'} - \delta\mid \mbf{f}_{\mcl{C} \cup \{\mbf{x}\}}, \epsilon_{\mbf{x}})\ \text{d}\epsilon_{\mbf{x}}\\
    &= \int_{-\infty}^{\infty}
        p(\epsilon_{\mbf{x}})
        \prod_{\mbf{x}' \in \mcl{C}\setminus \{\mbf{x}\}} p(\epsilon_{\mbf{x}'} \le \epsilon_{\mbf{x}} + f_{\mbf{x}} - f_{\mbf{x}'} - \delta\mid  \mbf{f}_{\mcl{C} \cup \{\mbf{x}\}}, \epsilon_{\mbf{x}})\ \text{d}\epsilon_{\mbf{x}}\\
    &= \int_{-\infty}^{\infty}
        e^{-\epsilon_{\mbf{x}}} e^{- e^{-\epsilon_{\mbf{x}}}}
        \prod_{\mbf{x}' \in \mcl{C}\setminus \{\mbf{x}\}} e^{-e^{-(\epsilon_{\mbf{x}} + f_{\mbf{x}} - f_{\mbf{x}'} - \delta)}}\ \text{d}\epsilon_{\mbf{x}}\\
    &= \int_{-\infty}^{\infty}
        e^{-\epsilon_{\mbf{x}}} e^{- e^{-\epsilon_{\mbf{x}}}}
        e^{-\sum_{\mbf{x}' \in \mcl{C}\setminus \{\mbf{x}\}} e^{-(\epsilon_{\mbf{x}} + f_{\mbf{x}} - f_{\mbf{x}'} - \delta)}}\ \text{d}\epsilon_{\mbf{x}}\\
    &= \int_{-\infty}^{\infty}
        e^{-\epsilon_{\mbf{x}}} e^{- e^{-\epsilon_{\mbf{x}}}}
        \left(e^{-e^{-\epsilon_{\mbf{x}}}}\right)^{\sum_{\mbf{x}' \in \mcl{C}\setminus \{\mbf{x}\}} e^{-(f_{\mbf{x}} - f_{\mbf{x}'} - \delta)}}\ \text{d}\epsilon_{\mbf{x}}\ .
\end{align*}
Let $t_{\mbf{x}} = e^{-e^{-\epsilon_{\mbf{x}}}}$, then $\text{d}t_{\mbf{x}} = e^{-\epsilon_{\mbf{x}}} e^{- e^{-\epsilon_{\mbf{x}}}}\ \text{d}\epsilon_{\mbf{x}}$, and 
\begin{align*}
    t_{\mbf{x}} = \begin{cases}
        0 &\text{ if } \epsilon_{\mbf{x}} = -\infty\ ,\\
        1 &\text{ if } \epsilon_{\mbf{x}} = \infty\ .
    \end{cases}
\end{align*}
Therefore, by change of variable $\epsilon_{\mbf{x}}$ to $t_{\mbf{x}}$,
\begin{align*}
&p(\mbf{x} \succ \mcl{C} \setminus \{\mbf{x}\}\mid  \mbf{f}_{\mcl{C} \cup \{\mbf{x}\}}; \delta) = \int_0^1 t_{\mbf{x}}^{\sum_{\mbf{x}' \in \mcl{C}\setminus \{\mbf{x}\}} e^{-(f_{\mbf{x}} - f_{\mbf{x}'} - \delta)}}\ \text{d}t_{\mbf{x}}\\
    &= \frac{t_{\mbf{x}}^{1 + \sum_{\mbf{x}' \in \mcl{C}\setminus \{\mbf{x}\}} e^{-(f_{\mbf{x}} - f_{\mbf{x}'} - \delta)}}}{
        1+ \sum_{\mbf{x}' \in \mcl{C}\setminus \{\mbf{x}\}} e^{-(f_{\mbf{x}} - f_{\mbf{x}'} - \delta)}
    } \Bigg|_0^1 = \frac{1}{
        1+ \sum_{\mbf{x}' \in \mcl{C}\setminus \{\mbf{x}\}} e^{-(f_{\mbf{x}} - f_{\mbf{x}'} - \delta)}
    } = \frac{e^{f_{\mbf{x}}}}{
        e^{f_{\mbf{x}}}
        + \sum_{\mbf{x}' \in \mcl{C}\setminus \{\mbf{x}\}} e^{f_{\mbf{x}'} + \delta}
    }\ .
\end{align*}

\section{Ranking and Pairwise Preference}
\label{app:rankpairwise}
In this section, we show that it is not trivial to map the probability of a ranking to probabilities of pairwise preferences. Hence, the work of \citet{gonzalez2017preferential} is not easily generalizable to rankings.
For simplicity, we do not consider ties in this section.
We show that the following two trivial mappings violate an axiom of probability. Let us consider the probability of a ranking of $3$ inputs $\mbf{x}_0$, $\mbf{x}_1$, and $\mbf{x}_2$. 
If we only use pairwise preferences to express the probability of ranking among these inputs, then a straightforward approach is to express it as a product of the probabilities of all pairwise preferences in the ranking, e.g.,
\begin{align*}
p(\mbf{x}_0 \succ \mbf{x}_1 \succ \mbf{x}_2) = p(\mbf{x}_0 \succ \mbf{x}_1)\ p(\mbf{x}_1 \succ \mbf{x}_2)\ p(\mbf{x}_0 \succ \mbf{x}_2)\ .
\end{align*}
Using this expression, we can compute the probability of $\mbf{x}_0 \succ \mbf{x}_1$ by marginalizing over all possible rankings:
\begin{align}
&p(\mbf{x}_0 \succ \mbf{x}_1) 
    = p(\mbf{x}_0 \succ \mbf{x}_1 \succ \mbf{x}_2) + p(\mbf{x}_0 \succ \mbf{x}_2 \succ \mbf{x}_1) + p(\mbf{x}_2 \succ \mbf{x}_0 \succ \mbf{x}_1)\label{eq:firstry}\\
    &= p(\mbf{x}_0 \succ \mbf{x}_1)\ p(\mbf{x}_1 \succ \mbf{x}_2)\ p(\mbf{x}_0 \succ \mbf{x}_2) 
        + p(\mbf{x}_0 \succ \mbf{x}_2)\  p(\mbf{x}_2 \succ \mbf{x}_1)\  p(\mbf{x}_0 \succ \mbf{x}_1)\nonumber\\
        &\quad\quad+ p(\mbf{x}_2 \succ \mbf{x}_0)\ p(\mbf{x}_0 \succ \mbf{x}_1)\ p(\mbf{x}_2 \succ \mbf{x}_1)\nonumber\\
    &= p(\mbf{x}_0 \succ \mbf{x}_1)
    \left\{
        p(\mbf{x}_0 \succ \mbf{x}_2)
        \left[
            p(\mbf{x}_1 \succ \mbf{x}_2)
            +  p(\mbf{x}_2 \succ \mbf{x}_1)
        \right]
        + p(\mbf{x}_2 \succ \mbf{x}_0)\  p(\mbf{x}_2 \succ \mbf{x}_1)
        \right\}\nonumber\\
    &= p(\mbf{x}_0 \succ \mbf{x}_1)
    \left\{
        p(\mbf{x}_0 \succ \mbf{x}_2)
        + p(\mbf{x}_2 \succ \mbf{x}_0)\  p(\mbf{x}_2 \succ \mbf{x}_1)
    \right\}\nonumber\\
    &= p(\mbf{x}_0 \succ \mbf{x}_1)
    \left\{
        1 - p(\mbf{x}_2 \succ \mbf{x}_0)
        + p(\mbf{x}_2 \succ \mbf{x}_0)\  p(\mbf{x}_2 \succ \mbf{x}_1)
    \right\}\nonumber\\
    &= p(\mbf{x}_0 \succ \mbf{x}_1)
    \left\{
        1 + p(\mbf{x}_2 \succ \mbf{x}_0) \left[p(\mbf{x}_2 \succ \mbf{x}_1) - 1\right]
    \right\}\ .\nonumber
\end{align}
Hence,
\begin{equation*}
    p(\mbf{x}_0 \succ \mbf{x}_1)\ 
        p(\mbf{x}_2 \succ \mbf{x}_0) \left[p(\mbf{x}_2 \succ \mbf{x}_1) - 1\right] = 0\ .
\end{equation*}
Therefore, if $p(\mbf{x}_0 \succ \mbf{x}_1) > 0$, $p(\mbf{x}_2 \succ \mbf{x}_0) > 0$, and $p(\mbf{x}_2 \succ \mbf{x}_1) < 1$, then \eqref{eq:firstry} does not hold. In other words, the $\sigma$-additivity axiom of probability is violated.

On the other hand, one can reason that $\mbf{x}_0 \succ \mbf{x}_1$ and $\mbf{x}_1 \succ \mbf{x}_2$ may imply $\mbf{x}_0 \succ \mbf{x}_2$. So, the probability of a ranking can be expressed as a product of the probabilities of consecutive pairs in the ranking, e.g.,
\begin{align*}
p(\mbf{x}_0 \succ \mbf{x}_1 \succ \mbf{x}_2) = p(\mbf{x}_0 \succ \mbf{x}_1)\ p(\mbf{x}_1 \succ \mbf{x}_2)\ .
\end{align*}
Let $a$, $b$, and $c$ denote $p(\mbf{x}_0 \succ \mbf{x}_1)$, $p(\mbf{x}_1 \succ \mbf{x}_2)$, and $p(\mbf{x}_2 \succ \mbf{x}_0)$, respectively (i.e., $a,b,c \in [0,1]$). The $\sigma$-additivity axiom of probability leads to the following expression:
\begin{align}
p(\mbf{x}_0 \succ \mbf{x}_1) 
    &= p(\mbf{x}_0 \succ \mbf{x}_1 \succ \mbf{x}_2) + p(\mbf{x}_0 \succ \mbf{x}_2 \succ \mbf{x}_1) + p(\mbf{x}_2 \succ \mbf{x}_0 \succ \mbf{x}_1)\label{eq:secondtry}\\
a &= ab + (1-c)(1-b) + ca\nonumber\\
a &= ab + bc + ca - b - c + 1\nonumber\\
a + b - ac - bc + c - 1 &= ab\nonumber\\
(a + b - 1)(1 - c) &= ab\ .
\label{eq:rankpair2}
\end{align}
Hence, if $ab > 0$ and $a + b < 1$ (e.g., $a = 0.1$ and $b = 0.2$), then there is no value of $c \in [0,1]$ satisfying \eqref{eq:rankpair2} (which requires $c > 1$). Thus, if $a = 0.1$ and $b = 0.2$, then \eqref{eq:secondtry} does not hold. In other words, the $\sigma$-additivity axiom of probability is violated.

\section{Complication When Ties Exist for $k > 1$}
\label{app:tieklarge1}

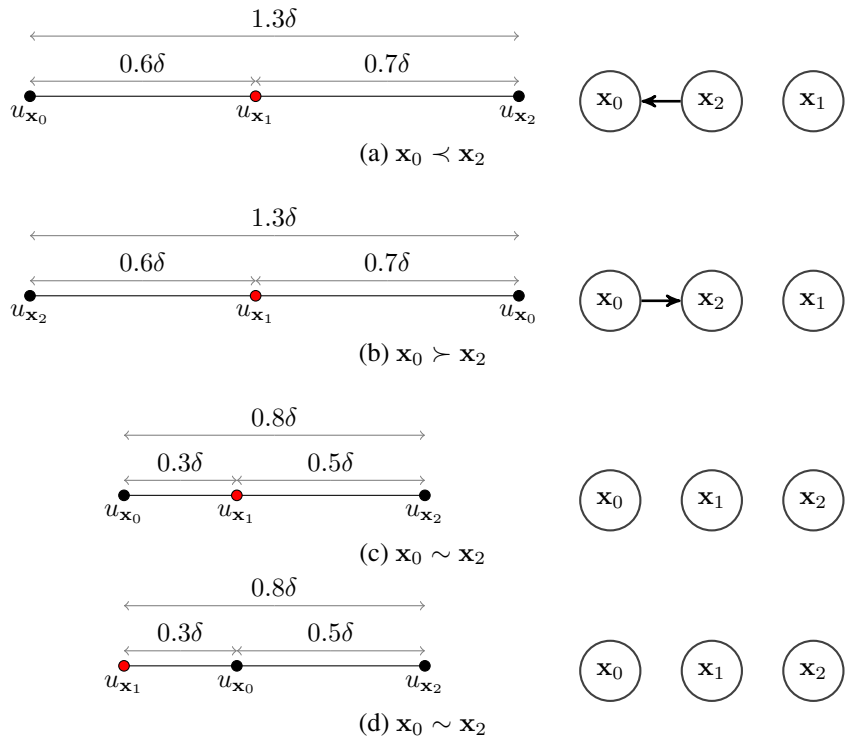
\begin{figure}[h]
\centering
\begin{tabular}{cc}
    \begin{tikzpicture}
    \draw[gray,<-] (0,0.2) -- (1.5,0.2); 
    \draw[black] (1.5,0.2) node[above,align=center]{$0.6\delta$};
    \draw[gray,->] (1.5,0.2) -- (3,0.2);
    \draw[gray,<-] (3,0.2) -- (4.75,0.2); 
    \draw[black] (4.75,0.2) node[above,align=center]{$0.7\delta$};
    \draw[gray,->] (4.75,0.2) -- (6.5,0.2);
    \draw[gray,<-] (0,0.8) -- (3.25,0.8); 
    \draw[black] (3.25,0.8) node[above,align=center]{$1.3\delta$};
    \draw[gray,->] (3.25,0.8) -- (6.5,0.8);

    \draw (0,0) node[align=center,   below] {$u_{\mbf{x}_0}$} -- (3,0) node[align=center, below] {$u_{\mbf{x}_1}$} -- (6.5,0) node[align=center,  below] {$u_{\mbf{x}_2}$};  
    \draw[black,fill=black] (0,0) circle (2pt);
    \draw[black,fill=red] (3,0) circle (2pt);
    \draw[black,fill=black] (6.5,0) circle (2pt);
    \end{tikzpicture}
&
    \begin{tikzpicture}[node distance=1.35cm,>=stealth',bend angle=45,auto]
      \tikzstyle{rv}=[circle,thick,draw=black!75,fill=white,minimum size=0.8cm]
    
    	\node[rv] (x0) {$\mbf{x}_0$};
    	\node[rv, right of=x0] (x2) {$\mbf{x}_2$};
    	\node[rv, right of=x2] (x1) {$\mbf{x}_1$};
    	
    	\draw[->,draw=black, line width=1]
    		(x2) edge node{} (x0);
    \end{tikzpicture}
\\
\multicolumn{2}{c}{(a) $\mbf{x}_0 \prec \mbf{x}_2$}
\\
\\

    \begin{tikzpicture}
    \draw[gray,<-] (0,0.2) -- (1.5,0.2); 
    \draw[black] (1.5,0.2) node[above,align=center]{$0.6\delta$};
    \draw[gray,->] (1.5,0.2) -- (3,0.2);
    \draw[gray,<-] (3,0.2) -- (4.75,0.2);
    \draw[black] (4.75,0.2) node[above,align=center]{$0.7\delta$};
    \draw[gray,->] (4.75,0.2) -- (6.5,0.2);
    \draw[gray,<-] (0,0.8) -- (3.25,0.8);
    \draw[black] (3.25,0.8) node[above,align=center]{$1.3\delta$};
    \draw[gray,->] (3.25,0.8) -- (6.5,0.8);

    \draw (0,0) node[align=center,   below] {$u_{\mbf{x}_2}$} -- (3,0) node[align=center, below] {$u_{\mbf{x}_1}$} -- (6.5,0) node[align=center,  below] {$u_{\mbf{x}_0}$};  
    \draw[black,fill=black] (0,0) circle (2pt);
    \draw[black,fill=red] (3,0) circle (2pt);
    \draw[black,fill=black] (6.5,0) circle (2pt);
    \end{tikzpicture}
&
    \begin{tikzpicture}[node distance=1.35cm,>=stealth',bend angle=45,auto]
      \tikzstyle{rv}=[circle,thick,draw=black!75,fill=white,minimum size=0.8cm]
    
    	\node[rv] (x0) {$\mbf{x}_0$};
    	\node[rv, right of=x0] (x2) {$\mbf{x}_2$};
    	\node[rv, right of=x2] (x1) {$\mbf{x}_1$};
    	
    	\draw[->,draw=black, line width=1]
    		(x0) edge node{} (x2);
    \end{tikzpicture}
\\
\multicolumn{2}{c}{(b) $\mbf{x}_0 \succ \mbf{x}_2$}
\\
\\
    \begin{tikzpicture}
    \draw[gray,<-] (0,0.2) -- (0.75,0.2); 
        \draw[black] (0.75,0.2) node[above,align=center]{$0.3\delta$};
    \draw[gray,->] (0.75,0.2) -- (1.5,0.2);
    \draw[gray,<-] (1.5,0.2) -- (2.75,0.2); 
        \draw[black] (2.75,0.2) node[above,align=center]{$0.5\delta$};
    \draw[gray,->] (2.75,0.2) -- (4,0.2);
    \draw[gray,<-] (0,0.8) -- (2,0.8); 
        \draw[black] (2,0.8) node[above,align=center]{$0.8\delta$};
    \draw[gray,->] (2,0.8) -- (4,0.8);
    
    \draw (0,0) node[align=center,   below] {$u_{\mbf{x}_0}$} -- (1.5,0) node[align=center, below] {$u_{\mbf{x}_1}$} -- (4,0) node[align=center,  below] {$u_{\mbf{x}_2}$};  
    \draw[black,fill=black] (0,0) circle (2pt);
    \draw[black,fill=red] (1.5,0) circle (2pt);
    \draw[black,fill=black] (4,0) circle (2pt);
    \end{tikzpicture}
&
    \begin{tikzpicture}[node distance=1.35cm,>=stealth',bend angle=45,auto]
      \tikzstyle{rv}=[circle,thick,draw=black!75,fill=white,minimum size=0.8cm]
    
    	\node[rv] (x0) {$\mbf{x}_0$};
    	\node[rv, right of=x0] (x1) {$\mbf{x}_1$};
    	\node[rv, right of=x1] (x2) {$\mbf{x}_2$};
    \end{tikzpicture}
\\
\multicolumn{2}{c}{(c) $\mbf{x}_0 \sim \mbf{x}_2$}
\\
    \begin{tikzpicture}
    \draw[gray,<-] (0,0.2) -- (0.75,0.2); 
        \draw[black] (0.75,0.2) node[above,align=center]{$0.3\delta$};
    \draw[gray,->] (0.75,0.2) -- (1.5,0.2);
    \draw[gray,<-] (1.5,0.2) -- (2.75,0.2); 
        \draw[black] (2.75,0.2) node[above,align=center]{$0.5\delta$};
    \draw[gray,->] (2.75,0.2) -- (4,0.2);
    \draw[gray,<-] (0,0.8) -- (2,0.8); 
        \draw[black] (2,0.8) node[above,align=center]{$0.8\delta$};
    \draw[gray,->] (2,0.8) -- (4,0.8);
    
    \draw (0,0) node[align=center,   below] {$u_{\mbf{x}_1}$} -- (1.5,0) node[align=center, below] {$u_{\mbf{x}_0}$} -- (4,0) node[align=center,  below] {$u_{\mbf{x}_2}$};  
    \draw[black,fill=red] (0,0) circle (2pt);
    \draw[black,fill=black] (1.5,0) circle (2pt);
    \draw[black,fill=black] (4,0) circle (2pt);
    \end{tikzpicture}
&
    \begin{tikzpicture}[node distance=1.35cm,>=stealth',bend angle=45,auto]
      \tikzstyle{rv}=[circle,thick,draw=black!75,fill=white,minimum size=0.8cm]
    
    	\node[rv] (x0) {$\mbf{x}_0$};
    	\node[rv, right of=x0] (x1) {$\mbf{x}_1$};
    	\node[rv, right of=x1] (x2) {$\mbf{x}_2$};
    \end{tikzpicture}
\\
\multicolumn{2}{c}{(d) $\mbf{x}_0 \sim \mbf{x}_2$}
\end{tabular}
\caption{Different preferences between $\mbf{x}_0$ and $\mbf{x}_2$ given $\mbf{x}_0 \sim \mbf{x}_1$ and $\mbf{x}_1 \sim \mbf{x}_2$. The graphs on the right column show the corresponding preference order relationship where arrows show the direction of preference and no connection means indifference/tie.}
\label{fig:diffpref}
\end{figure}

\begin{figure}
\centering
\begin{tabular}{c}
    \begin{tikzpicture}
    \draw[gray,<-] (0,0.2) -- (0.5,0.2); 
        \draw[black] (0.5,0.2) node[above,align=center]{$0.1\delta$};
    \draw[gray,->] (0.5,0.2) -- (1,0.2);
    
    \draw[gray,<-] (1,0.2) -- (4,0.2); 
        \draw[black] (4,0.2) node[above,align=center]{$0.6\delta$};
    \draw[gray,->] (4,0.2) -- (7,0.2);
    
    \draw[gray,<-] (7,0.2) -- (8,0.2);
        \draw[black] (8,0.2) node[above,align=center]{$0.2\delta$};
    \draw[gray,->] (8,0.2) -- (9,0.2);
    
    \draw[gray,<-] (9,0.2) -- (10.5,0.2);
        \draw[black] (10.5,0.2) node[above,align=center]{$0.3\delta$};
    \draw[gray,->] (10.5,0.2) -- (12,0.2);
    
    \draw[gray,<-] (1,0.8) -- (6.5,0.8);
        \draw[black] (6.5,0.8) node[above,align=center]{$1.1\delta$};
    \draw[gray,->] (6.5,0.8) -- (12,0.8);
    
    \draw (0,0) node[align=center,   below] {$u_{\mbf{x}_0}$} -- (1,0) node[align=center, below] {$u_{\mbf{x}_4}$} 
    -- (7,0) node[align=center,  below] {$u_{\mbf{x}_1}$}
    -- (9,0) node[align=center,  below] {$u_{\mbf{x}_3}$}
    -- (12,0) node[align=center,  below] {$u_{\mbf{x}_2}$};  
    \draw[black,fill=black] (0,0) circle (2pt);
    \draw[black,fill=black] (1,0) circle (2pt);
    \draw[black,fill=black] (7,0) circle (2pt);
    \draw[black,fill=black] (9,0) circle (2pt);
    \draw[black,fill=black] (12,0) circle (2pt);
    \end{tikzpicture}
\\
(a)
\\
\\
    \begin{tikzpicture}[node distance=1.35cm,>=stealth',bend angle=45,auto]
      \tikzstyle{rv}=[circle,thick,draw=black!75,fill=white,minimum size=0.8cm]
    
    	\node[rv] (x2) {$\mbf{x}_2$};
    	\node[rv, below of=x2, xshift=-0.65cm] (x0) {$\mbf{x}_0$};
    	\node[rv, below of=x2, xshift=0.65cm] (x4) {$\mbf{x}_4$};
    	\node[rv, right of=x2] (x1) {$\mbf{x}_1$};
    	\node[rv, right of=x1] (x3) {$\mbf{x}_3$};
    	
    	\draw[->,draw=black, line width=1]
    		(x2) edge node{} (x0)
    		(x2) edge node{} (x4);
    
    \end{tikzpicture}
\\
(b)
\end{tabular}
\caption{An example of a (partial) ranking of $5$ inputs. The graph in (b) shows the preference order relationship where arrows show the direction of preference and no connection means indifference/tie.}
\label{fig:diffpref5}
\end{figure}
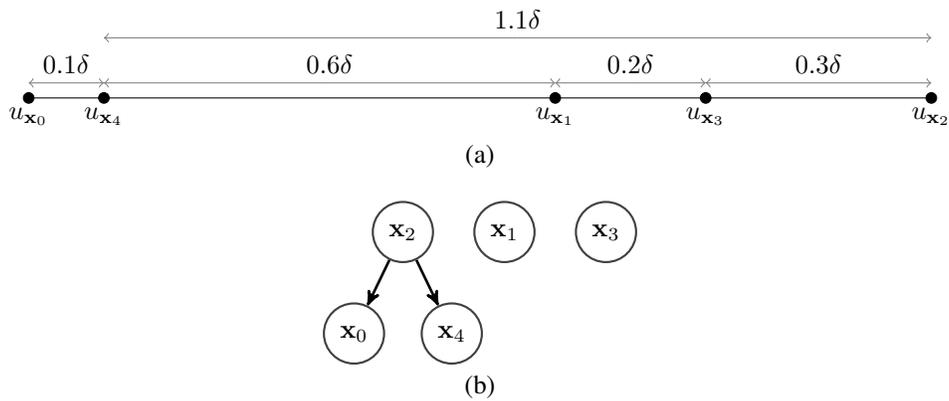

To illustrate the complication in dealing with tie observation for $k > 1$ due to the partial order of preference, we consider the preference among $3$ different inputs: $\mbf{x}_0$, $\mbf{x}_1$, and $\mbf{x}_2$. Let us denote $\mbf{x} \sim \mbf{x}'$ as a tie preference between $\mbf{x}$ and $\mbf{x}'$. The observation $\mbf{x}_0 \sim \mbf{x}_1 \sim \mbf{x}_2$ (i.e., $\mbf{x}_0 \sim \mbf{x}_1$ and $\mbf{x}_1 \sim \mbf{x}_2$) can lead to $3$ different preferences between $\mbf{x}_0$ and $\mbf{x}_2$, as shown in Fig.~\ref{fig:diffpref}. 
The difficulty is due to the fact that transitivity does not hold for ties, i.e., $\mbf{x}_0 \sim \mbf{x}_1$ and $\mbf{x}_1 \sim \mbf{x}_2$ do not imply $\mbf{x}_0 \sim \mbf{x}_2$ (Fig.~\ref{fig:diffpref}a).
The interpretation of the observation becomes more complicated when we would like to fully describe the preference (with tie) among $|\mcl{C}| > 3$ inputs.
For example, it is possible that $\mbf{x}_0 \sim \mbf{x}_1 \sim \mbf{x}_2 \sim \mbf{x}_3 \sim \mbf{x}_4$ and $\mbf{x}_0 \sim \mbf{x}_4$, but $\mbf{x}_0 \prec \mbf{x}_2 \succ \mbf{x}_4$ (Fig.~\ref{fig:diffpref5}).
Thus, the probability of an observation involving ties cannot be computed as straightforwardly as \eqref{eq:tieprob}.
We decide to leave this scenario for future investigation and focus on $k = 1$ when tie exists so that it does not unnecessarily complicate our exposition of the BO model and the acquisition function.

\section{Additional BO Experiments with Ties}
\label{app:notie}

\begin{figure}[h!]
\centering
\begin{tabular}{@{}ccc@{}}
\includegraphics[height=0.2\textwidth]{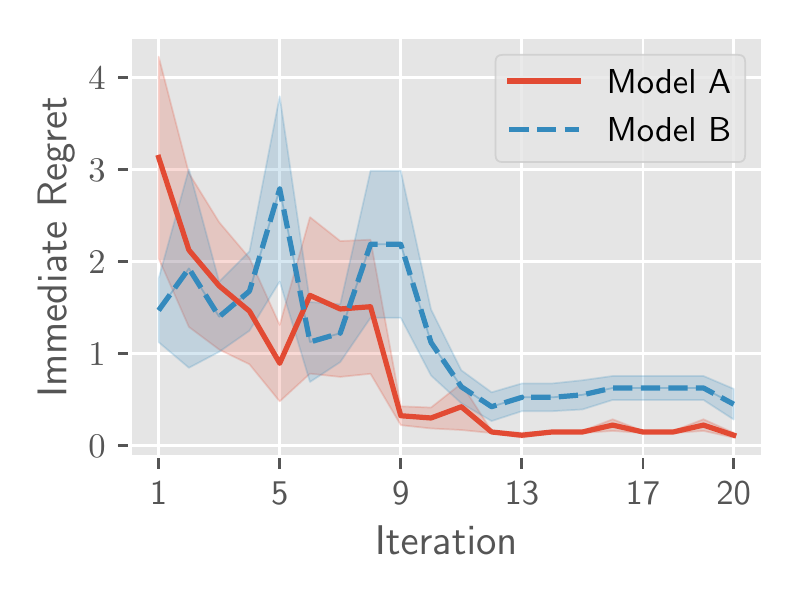}
&
\includegraphics[height=0.2\textwidth]{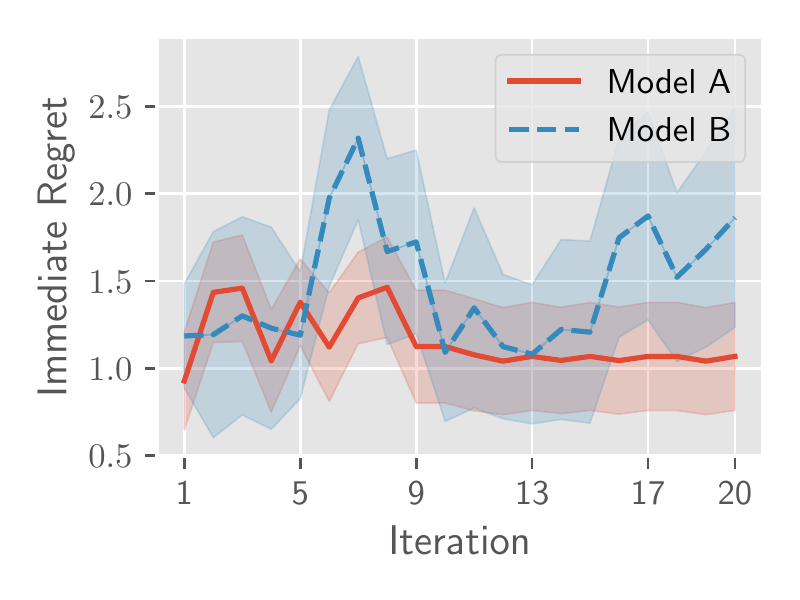}
&
\includegraphics[height=0.2\textwidth]{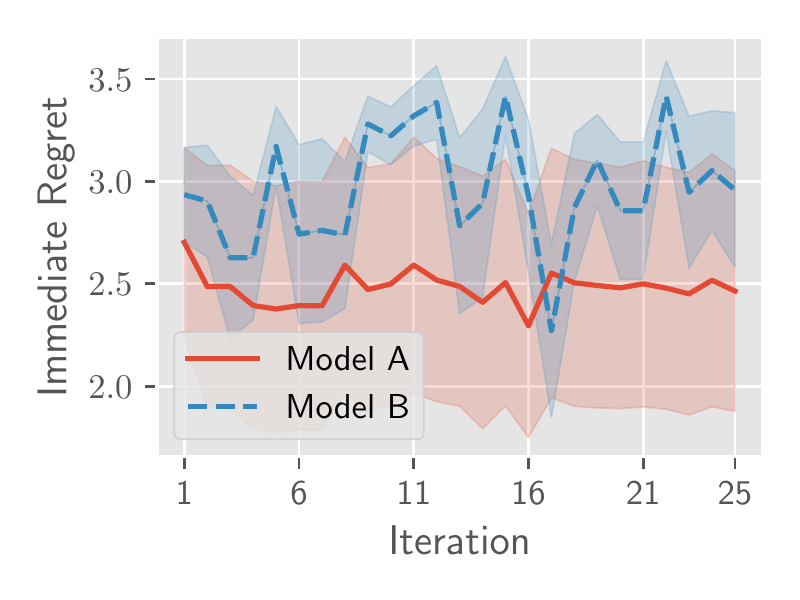}
\\
(a) Forrester function
&
(b) SHC function
&
(c) Hartmann function
\end{tabular}
\caption{Plots of immediate regrets for experiments with pairwise preferences and ties. MPES is used with model A that handles ties and model B that cannot handle ties; the tie observation is converted to a random strict pairwise preference for model B.}
\label{fig:convertie}
\end{figure}

To highlight the importance of modeling ties in Sec.~\ref{subsec:tie}, we consider BO problems with tie observations and compare the performance of MPES with $2$ models: (model A) the model is capable of handling ties (as specified in Sec.~\ref{subsec:tie}) and (model B) the model cannot handle ties.
In other words, while model A accepts both strict pairwise preferences and ties, model B only accepts strict preferences, i.e., $\delta = 0$. Therefore, in order to train model B with ties in the observation of the BO problems, we convert ties to random strict preferences.
For example, if the observation is a tie $\mbf{x} \sim \mbf{x}'$, we randomly convert this tie to either $\mbf{x} \succ \mbf{x}'$ or $\mbf{x} \prec \mbf{x}'$ with equal probabilities.
The rationale behind this conversion is that when we are indifferent about a number of inputs in a set $\mcl{C}$ (a tie) and yet, we are forced to choose an input as the most preferred (since the model cannot handle ties), we choose a random input in $\mcl{C}$ because the preferences of all inputs in $\mcl{C}$ are the same to us.

These BO problems are about optimizing synthetic benchmark functions: Forrester, six-hump camel (SHC), and $3$-D Hartmann functions. The threshold $\delta$ to generate observations (including ties) is set to $2$. We set $k = 1$ and $|\mcl{C}| = 2$, i.e., the possible observations are strict pairwise preferences and ties between $2$ inputs. The threshold $\delta$ is unknown and is learned from the observation for model A, while $\delta = 0$ for model B as it represents a model that cannot handle ties.
We repeat each experiment $5$ times to plot the average and the standard error of the immediate regret in Fig.~\ref{fig:convertie}. It can be observed that by modeling ties, the BO performance of MPES with model A is improved as compared to that with model B which can only handle strict preferences.
This is because randomly assigning the strict preference due to ties/indifference can lead to an inaccurate update of the posterior belief of the objective function.
As a result, it is important to model the real-world tie observation such as in Sec.~\ref{subsec:tie} to achieve a competitive BO performance.

\end{document}